\pdfoutput=1

\documentclass[11pt]{article}

\usepackage[preprint]{acl}

\usepackage{times}
\usepackage{amsmath}
\usepackage{latexsym}
\usepackage{todonotes}
\usepackage{caption}
\usepackage{subcaption}
\usepackage{booktabs}
\usepackage{comment}
\usepackage{multirow}
\usepackage{verbatim}
\usepackage{tcolorbox}
\usepackage{tabularx}
\usepackage{url}
\usepackage{makecell}

\usepackage[T1]{fontenc}

\usepackage[utf8]{inputenc}

\usepackage{microtype}

\usepackage{inconsolata}

\usepackage{graphicx}

\title{DogeRM: Equipping Reward Models with Domain Knowledge\\through Model Merging}

\author{Tzu-Han Lin \quad Chen-An Li \quad Hung-yi Lee \quad Yun-Nung Chen \\
National Taiwan University, Taipei, Taiwan \\
\texttt{\{r12944034,r13942069,hungyilee\}@ntu.edu.tw} \quad \texttt{y.v.chen@ieee.org}
}

\begin{document}

\maketitle

\begin{abstract}

Reinforcement learning from human feedback (RLHF) is a popular strategy for aligning large language models (LLMs) with desired behaviors. Reward modeling is a crucial step in RLHF. However, collecting paired preference data for training reward models is often costly and time-consuming, especially for domain-specific preferences requiring expert annotation. To address this challenge, we propose the \textbf{Do}main knowled\textbf{ge} merged \textbf{R}eward \textbf{M}odel (DogeRM), a novel framework that integrates domain-specific knowledge into a general reward model by model merging. The experiments demonstrate that DogeRM enhances performance across different benchmarks and provide a detailed analysis showcasing the effects of model merging, showing the great potential of facilitating model alignment.\footnote{The source code and trained models are released at \\\url{https://github.com/MiuLab/DogeRM}.}

\end{abstract}

\section{Introduction}

Modern large language models (LLMs), such as GPT-4~\citep{achiam2023gpt} and Gemini~\citep{team2023gemini}, showcase impressive capabilities across various tasks~\citep{eval-harness, open-llm-leaderboard}, with aligning their behavior with human preferences. Reinforcement learning from human feedback (RLHF) is a prominent technique for enhancing the alignment of desired behaviors in LLMs~\citep{christiano2017deep,ziegler2020finetuning,instructgpt}. A key component of RLHF is its reward models (RMs), which assess entire sentences generated by policy models. The reward signals produced by these RMs are instrumental in adjusting the parameters of the policy models, thus directly impacting the policy models' effectiveness.

RMs are developed by training LLMs on \emph{paired} preference data to simulate human judgment~\citep{instructgpt}. This preference data consists of two responses to a given user input, accompanied by a human-assigned label indicating which response is more preferred. However, gathering such preference data can be costly and time-consuming due to the requirement of human annotation~\citep{openaisummarize}. This challenge becomes more pronounced when handling domain-specific preference data, as it necessitates expertise from domain specialists.

\begin{figure}[t!]
    \centering    
    \includegraphics[width=\linewidth]{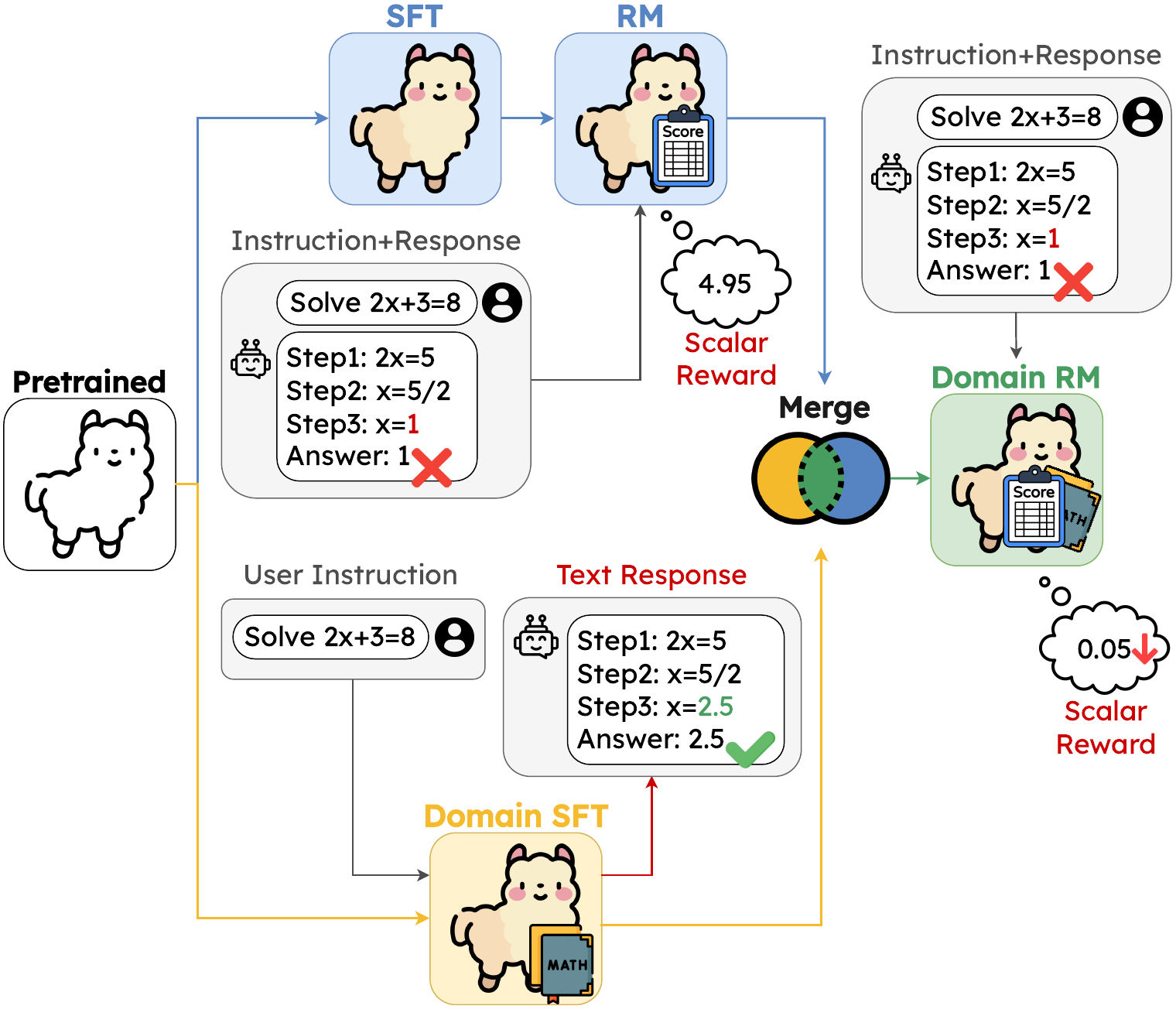}
    \caption{The framework of \textbf{DogeRM}, illustrating the merging of a general RM with a domain-specific LM to create a domain-specific RM. All icons used in this figure are sourced from \url{https://www.flaticon.com/}.}
    \label{fig:overview}
\end{figure}

Recent developments have demonstrated the effectiveness of model merging techniques in strategically integrating multiple domain-specific models into a multi-domain model without requiring additional training~\citep{wortsman2022model, ilharco2023editing}. Furthermore, domain-specific SFT data is relatively more accessible compared to preference data. Moreover, many high-quality domain-specific models are available on open-source platforms~\citep{wolf-etal-2020-transformers}, which can be directly employed in the merging process. This brings us to consider a novel approach: 
\textit{Is it possible to equip reward models with domain knowledge through merging with domain-specific language models?}

In this work, we propose \textbf{Do}main knowled\textbf{ge} merged \textbf{R}eward \textbf{M}odel (DogeRM), exploring the potential of merging a reward model trained on a general open-sourced preference dataset with a language model fine-tuned on domain-specific datasets, such as math and coding. An illustration of DogeRM is presented in Figure~\ref{fig:overview}. We evaluate DogeRM using RewardBench~\citep{lambert2024rewardbench}, Auto-J Eval~\citep{li2024generative} and Best-of-N sampling on  GSM8K~\citep{cobbe2021training} and MBPP~\citep{austin2021program}. Our results demonstrate that DogeRM improves performance and can be generalized to different model architectures. We also conduct a comprehensive analysis to demonstrate the impact of model merging.

\section{Related Work}

\paragraph{Reward Modeling}

RMs are crucial for aligning language models with human preferences, providing proxy rewards as training signals for policy models. Previous work has employed RL algorithms with RMs to guide language models towards human preferences in various NLP tasks~\citep{ziegler2020finetuning,openaisummarize,wu2021recursively,nakano2022webgpt,menick2022teaching} and instruction-following~\citep{instructgpt,ramamurthy2023is}. In RLHF literature, RMs evaluate the quality of instructions and responses based on criteria like helpfulness and harmlessness~\citep{bai2022training} or more fine-grained objectives~\citep{wu2023finegrained}.

Several open-source paired preference datasets are available for training RMs for RLHF, such as OpenAI Summarization~\citep{openaisummarize}, HH-RLHF~\citep{bai2022training}, SHP~\citep{pmlr-v162-ethayarajh22a}, Ultrafeedback~\citep{cui2024ultrafeedback}, PKU-SafeRLHF~\citep{pkualignment}, HelpSteer~\citep{wang-etal-2024-helpsteer}, Nectar~\citep{starling2023}, and UltraInteract~\citep{yuan2024advancing}. However, most datasets are not \emph{domain-specific}. To address this, our work focuses on merging RMs with domain-specific language models, aiming to equip RMs with domain knowledge.

\paragraph{Model Merging}

Model merging integrates multiple task-specific models into a single unified model without additional training. A straightforward approach involves averaging parameters from models fine-tuned from the same initial model~\citep{wortsman2022model}. Another method employs weighted averaging of model parameters~\citep{matena2022merging, jin2023dataless}.

Another innovative approach involves creating task vectors by subtracting the weights of a pre-trained model from those of the same model after fine-tuning for a specific task. This method showcases the flexibility and composability of these vectors through arithmetic operations~\citep{ilharco2023editing, yadav2024ties,huang-etal-2024-chat}.

Some recent work focused on model merging to align with user preferences. They interpolated model parameters fine-tuned on diverse rewards~\citep{rame2024rewarded, jang2023personalized, wang-etal-2024-arithmetic}, or merging RMs for combining different aspects of rewards~\citep{ramé2024warm}.

However, these methods still rely heavily on substantial domain-specific preference data to integrate domain knowledge. In contrast, our approach significantly reduces the need for such data by focusing on incorporating domain-specific knowledge into RMs through model merging.

\section{Methodology}

\subsection{Reward Modeling}

To train a reward model, we replace the decoding layer of a transformer-based pre-trained language model with a linear regression layer. This new layer projects the logits from the final transformer layer to a scalar, representing the reward of a given input.

Given an input prompt $x$, the chosen response $y_c$, and the rejected response $y_r$, we use the following loss function to optimize our reward model:
\begin{equation}
    \mathcal{L}_{\text{RM}} = -\log{[\sigma(r(x, y_c)) - \sigma(r(x, y_r))]}
\end{equation}
where $r(x, y_c)$ is the reward of chosen response, $r(x, y_r)$ is the reward of rejected response, and $\sigma(\cdot)$ is the logistic function.

\begin{table*}[hbtp]
    \centering
    \resizebox{\textwidth}{!}{
        \begin{tabular}{l c c c c c c c c c c}
        \toprule
        \multirow{4}{*}{\textbf{Model}} & \multicolumn{5}{c}{\textbf{Reward Bench}} & \multicolumn{3}{c}{\textbf{Auto-J Eval}} & \multicolumn{2}{c}{\textbf{Best-of-16}}\\
        \cmidrule(lr){2-6} \cmidrule(lr){7-9} \cmidrule(lr){10-11}
        & \multirow{2.5}{*}{Chat} & \multirow{2.5}{*}{Chat-Hard} & \multirow{2.5}{*}{Safety} & \multicolumn{2}{c}{Reasoning} & \multirow{2.5}{*}{Code} & \multirow{2.5}{*}{Math} & \multirow{2.5}{*}{Others} & \multirow{2.5}{*}{GSM8K} & \multirow{2.5}{*}{MBPP} \\
        \cmidrule(lr){5-6}
        & & & & Code & Math & & & & & \\
        \midrule
        \stepcounter{enumi}\makebox[0.5cm]{(\alph{enumi})} LLaMA-2 RM & 95.8 & 47.6 & \textbf{44.6} & 78.9 & 68.2 & 76.2 & 84.4 & 79.2 & 35.3 & \textbf{17.2} \\
        \midrule
        \stepcounter{enumi}\makebox[0.5cm]{(\alph{enumi})} FT on Auto-J Math & 94.7 & \textbf{48.5} & 44.4 & 79.1 & 68.7 & 76.2$^\dagger$ & \textbf{90.2}$^\dagger$ & 79.2$^\dagger$ & 35.2 & - \\
        \stepcounter{enumi}\makebox[0.5cm]{(\alph{enumi})} FT on Auto-J Code & 94.7 & 48.2 & 44.3 & 78.8 & 66.9 & \textbf{89.3}$^\dagger$ & 84.4$^\dagger$ & 79.4$^\dagger$ & - & \textbf{17.2} \\
        \midrule
        \stepcounter{enumi}\makebox[0.5cm]{(\alph{enumi})} Ours (+ MetaMath) & 95.8 & 44.5 & 43.5 & \textbf{85.7} & 79.6 & 79.8 & 87.5 & 79.3 & \textbf{40.7} & - \\
        \stepcounter{enumi}\makebox[0.5cm]{(\alph{enumi})} Ours (+ MAmmoTH) & \textbf{96.1} & 44.7 & 43.8 & 84.1 & \textbf{85.2} & 79.8 & 87.5 & \textbf{79.7} & 40.5 & - \\
        \stepcounter{enumi}\makebox[0.5cm]{(\alph{enumi})} Ours (+ Code Model) & \textbf{96.1} & 45.6 & 43.9 & 84.3 & 71.8 & 82.1 & 87.5 & \textbf{79.7} & - & \textbf{17.2} \\
        \bottomrule
        \end{tabular}
    }   

    \caption{Performance comparison across various benchmarks. Row (a) represents our base LLaMA-2 7B~\citep{touvron2023llama} reward model. Rows (b) and (c) show results after fine-tuning the LLaMA-2 RM using the test data from Auto-J Eval~\citep{li2024generative} Math and Code subsets, respectively. We use $\dagger$ to denote training accuracy, as these values are derived from benchmark testing data used during training. Rows (d) to (f) demonstrate the performance of LLaMA-2 RM when merged with MetaMath-7B~\citep{yu2024metamath}, MAmmoTH-7B~\citep{yue2024mammoth}, and the Code Model, each with a weight factor of $\lambda=0.35$.}
    \label{tab:benchmarks}
\end{table*}

\subsection{Model Merging}

Our proposed method merges the parameters of a supervised fine-tuned language model, denoted as $\theta^{\text{SFT}}$, with those of a reward model, $\theta^{\text{RM}}$, both initialized from the same pre-trained model $\theta$.

We divide $\theta^{\text{SFT}}$ into three disjoint parts:
\begin{equation}
    \theta^{\text{SFT}} = \{\theta^{\text{SFT}}_{\text{emb}}, \theta^{\text{SFT}}_{\text{trans}},  \theta^{\text{SFT}}_{\text{dec}}\}
\end{equation}
where $\theta^{\text{SFT}}_{\text{emb}}, \theta^{\text{SFT}}_{\text{trans}},\theta^{\text{SFT}}_{\text{dec}}$ represent the embedding, transformer, and decoding layers' parameters, respectively.

Similarly, we also divide $\theta^{\text{RM}}$ into three parts:
\begin{equation}
    \theta^{\text{RM}} = \{\theta^{\text{RM}}_{\text{emb}}, \theta^{\text{RM}}_{\text{trans}},  \theta^{\text{RM}}_{\text{reg}}\}
\end{equation}
where $\theta^{\text{RM}}_{\text{emb}}, \theta^{\text{RM}}_{\text{trans}},\theta^{\text{RM}}_{\text{reg}}$ denote the parameters for the embedding, transformer, and regression layer, respectively.

For embedding layer parameters, we apply a weighted average to common token embeddings: 
\begin{equation}
    \theta^{\text{MERGE}}_{\text{emb},t_i} = \lambda \cdot \theta^{\text{SFT}}_{\text{emb},t_i} + (1 - \lambda) \cdot \theta^{\text{RM}}_{\text{emb},t_i}
\end{equation}
where $t_i$ is a common token to both models, $\theta_{\text{emb},t_i}$ is the corresponding embedding, and $\lambda$ is a hyperparameter controlling the weight of the SFT parameters, ranging from 0 to 1.

As for the unshared tokens, we directly use the embedding from their corresponding source model.  
\begin{equation}
    \theta^{\text{MERGE}}_{\text{emb},t_i} = \begin{cases}
        \theta^{\text{SFT}}_{\text{emb},t_i} & \text{If $t_i$ is unique to SFT} \\
        \theta^{\text{RM}}_{\text{emb},t_i} & \text{If $t_i$ is unique to RM}
    \end{cases}
\end{equation}

For the transformer layers, we perform a weighted average directly since both models are initialized from the same pre-trained model: 
\begin{equation}
    \theta^{\text{MERGE}}_{\text{trans}} = \lambda \cdot \theta^{\text{SFT}}_{\text{trans}} + (1 - \lambda) \cdot \theta^{\text{RM}}_{\text{trans}}
\end{equation}

Finally, we derive the merged reward model $\theta^{\text{MERGE}}$ by combining $\theta^{\text{MERGE}}_{\text{emb}}$, $\theta^{\text{MERGE}}_{\text{trans}}$, and the reward model's regression layer $\theta^{\text{RM}}_{\text{reg}}$: 
\begin{equation}
    \theta^{\text{MERGE}} = \{\theta^{\text{MERGE}}_{\text{emb}}, \theta^{\text{MERGE}}_{\text{trans}}, \theta^{\text{RM}}_{\text{reg}}\}
\end{equation}

\section{Experiments}

\subsection{Experimental Setup}

\paragraph{Reward Model}

To fine-tune the backbone of our reward model, we utilize the 10k SFT split from Alpacafarm~\citep{dubois2023alpacafarm}. For reward modeling, we employ the UltraFeedback~\citep{cui2024ultrafeedback}. The details of training and these datasets are presented in Appendix~\ref{sec:appendix:training} and \ref{sec:appendix:dataset}, respectively.

\paragraph{Domain-Specific SFT}

For the math LLMs, we adopt the open-source models, MetaMath-7B~\citep{yu2024metamath} and MAmmoTH-7B~\citep{yue2024mammoth}, both of which are fine-tuned from LLaMA-2-7B. For code generation LLM, since we could not find open-source models with detailed training information, we use OSS-Instruct and Magicoder-Evol-Instruct~\citep{wei2024magicoder} to fine-tune LLaMA-2-7B ourselves. We refer to this model as the Code Model in the following sections. The details of code fine-tuning datasets and math models are presented in Appendix~\ref{sec:appendix:dataset}, \ref{sec:appendix:models}, and \ref{sec:appendix:code_model}. 

\subsection{Evaluation}

We evaluate the reward models using two benchmarks, RewardBench~\citep{lambert2024rewardbench} and Auto-J Eval~\citep{li2024generative}. These benchmarks provide paired instruction-completion data, with the preferred completion annotated as chosen and the other as rejected, using accuracy as the evaluation metric. We use the core set of RewardBench, focusing primarily on the reasoning category to evaluate the model's abilities in code and mathematical reasoning. For Auto-J Eval, we use pair-wise testing data and categorize the dataset into three categories: code, math, and others, following \citealt{yuan2024advancing}. Additionally, to further test our reward model's effectiveness in enhancing model performance through reranking, we conduct best-of-N sampling on zero-shot prompted responses from \texttt{llama-2-7B-chat} for GSM8K~\citep{cobbe2021training} and MBPP~\citep{austin2021program}. None of the models used in the experiment were trained on these data sources. Details about the models and datasets are provided in Appendix~\ref{sec:appendix:dataset} and \ref{sec:appendix:models}, while the hyperparameters used for best-of-N sampling are outlined in Appendix~\ref{sec:appendix:bon_hyperparameter}.

Additionally, in DogeRM, determining an appropriate weight factor $\lambda$ depends on a small in-domain validation set. This raises an important question: \textit{Can fine-tuning the reward model on this small dataset match or even exceed the performance of our method in the target domain?} To investigate this, we performed continuous fine-tuning of our LLaMA-2 RM using test data from Auto-J Eval and evaluated the newly fine-tuned RM on the remaining benchmarks to assess its effectiveness.

Lastly, We present results using a weight factor of $\lambda=0.35$ for the main findings, with an analysis of the impact of $\lambda$ detailed in section~\ref{subsec:analysis} and results for different values of $\lambda$ presented in Appendix~\ref{sec:appendix:full}.

\subsection{Results}

\paragraph{RM Benchmarks}

The main results of RM benchmarks are shown in Table~\ref{tab:benchmarks}. 
Merging the LLaMA-2 RM with MetaMath-7B (Row (d)) and MAmmoTH-7B (Row (e)) improves math performance on RewardBench by 11.4\% and 17\%, respectively, and coding performance by 5.2\% and 5.8\%, respectively. Similar enhancements are seen on Auto-J Eval, with gains in both math and coding. Merging our LLaMA-2 RM with the Code Model (Row (f)) further improves coding performance on RewardBench and Auto-J Eval by 5.4\% and 6\%, respectively, along with noticeable improvements in math performance on both benchmarks. Although DogeRM enhances performance in the reasoning domain, there is no significant degradation in other domains. The specific role of domain knowledge is evident, as merging with the math model leads to greater improvements in the math domain than merging with the Code model, and vice versa.

\paragraph{Best-of-N Sampling}

\begin{figure}[t!]
     \centering
     \begin{subfigure}[t]{0.493\linewidth}
         \centering
         \includegraphics[width=\textwidth]{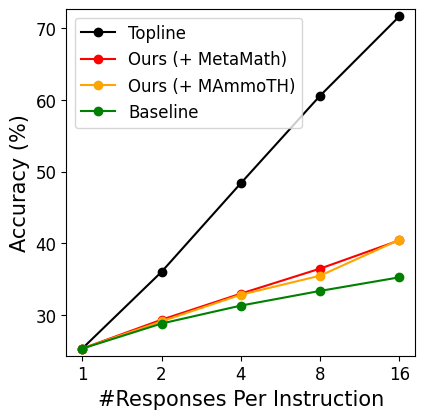}
         \caption{+ MetaMath/MAmmoTH on GSM8K.}
         \label{fig:bon_gsm8k}
     \end{subfigure}
     \hfill
     \begin{subfigure}[t]{0.493\linewidth}
         \centering
         \includegraphics[width=\textwidth]{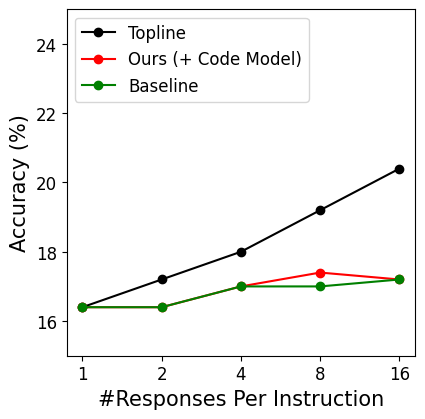}
         \caption{+ Code Model on MBPP.}
         \label{fig:bon_mbpp}
     \end{subfigure}

        \caption{Best-of-N results. Merging with domain-specific models improves reranking accuracy. Topline: Pass@N, the probability of obtaining at least one correct solution out of N responses. Baseline: LLaMA-2 RM.}
        \label{fig:bon_results}
\end{figure}

Figure~\ref{fig:bon_results} and Table~\ref{tab:benchmarks} show the results, with accuracy improvements on GSM8K. At the best-of-16 setting, merging with math models (Rows (d) and (e)) improves GSM8K by 5\%, while merging with the Code Model (Row (f)) maintains performance on MBPP without degradation.   We attribute the modest improvement on MBPP to the low upper bound of reranking performance (indicated by the black line in Figure~\ref{fig:bon_mbpp}), which constrains the potential gains from reranking in this task.

\paragraph{Fine-tuning on Small Validation Dataset}
Rows (b) and (c) of Table~\ref{tab:benchmarks} show the results of fine-tuning our LLaMA-2 RM on the Auto-J Eval Math and Code test subsets, respectively. While fine-tuning improved performance on Auto-J Eval, it did not generalize well to other benchmarks. In contrast, using these datasets as a validation set to determine an appropriate $\lambda$ for merging resulted in better overall performance. For a detailed analysis of $\lambda$'s impact across different benchmarks, see Appendix~\ref{sec:appendix:full}.

\subsection{Analysis}
\label{subsec:analysis}

\paragraph{Effect of Weight Factor $\lambda$}

\begin{figure}[t!]
     \centering
     \begin{subfigure}[h]{0.493\linewidth}
         \centering
         \includegraphics[width=\textwidth]{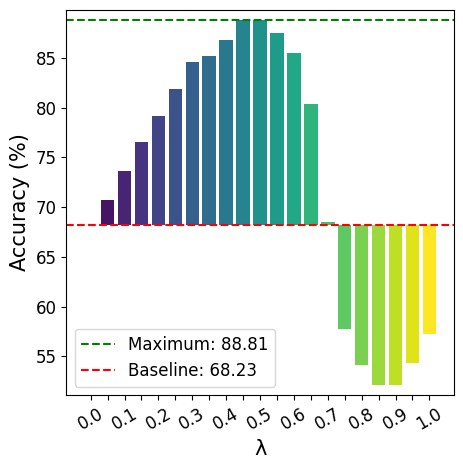}
         \caption{+ MAmmoTH on RewardBench math subset.}
         \label{fig:rb_mathprm}
     \end{subfigure}
     \hfill
     \begin{subfigure}[h]{0.493\linewidth}
         \centering
         \includegraphics[width=\textwidth]{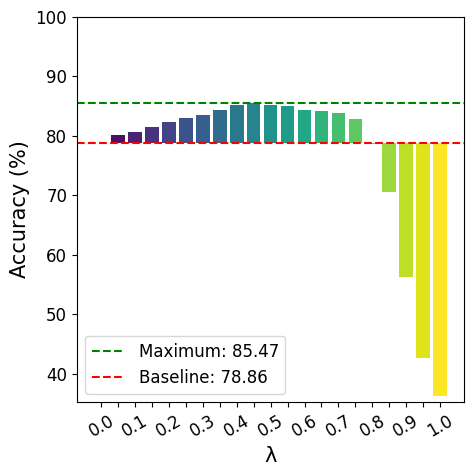}
         \caption{+ Code Model on RewardBench code subset.}
         \label{fig:rb_hep}
     \end{subfigure}
     \begin{subfigure}[h]{0.493\linewidth}
         \centering
         \includegraphics[width=\textwidth]{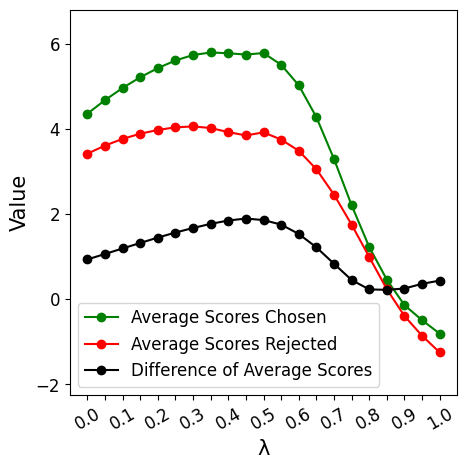}
         \caption{+ MAmmoTH on RewardBench math subset.}
         \label{fig:sd_mathprm}
     \end{subfigure}
     \hfill
     \begin{subfigure}[h]{0.493\linewidth}
         \centering
         \includegraphics[width=\textwidth]{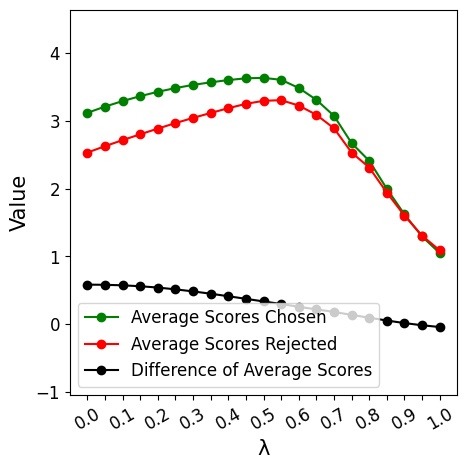}
         \caption{+ Code Model on RewardBench code subset.}
         \label{fig:sd_hep}
     \end{subfigure}

        \caption{The impact of different value of $\lambda$ on RewardBench math and code subsets. (a)(b): Accuracy; (c)(d): Reward difference between chosen and rejected prompts.}
        \label{fig:effect_lambda}
\end{figure}

To further investigate how the weight factor $\lambda$ affects our method's performance, we test various values of $\lambda$ ranging from 0 to 1 in increments of 0.05, observing the performance changes across these values on RewardBench. Figure~\ref{fig:effect_lambda} shows that performance degrades when $\lambda$ is large. We suggest setting $\lambda$ between 0.2 and 0.5 to achieve better results.

\paragraph{Reward Differences}

We delve deeper into how model merging affects the output of reward models by examining the value of the reward signals corresponding to chosen and rejected prompts in RewardBench. Figure~\ref{fig:effect_lambda} illustrates the distribution before and after merging. In the math subset, we notice that the difference in reward scores between the chosen and rejected prompts initially increases and then decreases as $\lambda$ varies from 0 to 1. Conversely, in the code subset, this difference consistently decreases. We hypothesize that this discrepancy arises because the original reward model inherently excels in the code subset.

\paragraph{Generalizability}

To test the adaptability of DogeRM to different model architectures, we use an open-source Mistral-based~\citep{jiang2023mistral} RM~\citep{mistralrm} merging with Misrtral-based MAmmoTH2-7B-Plus~\citep{yue2024mammoth2}. Details of these models are presented in Appendix~\ref{sec:appendix:models}. The results for $\lambda=0.35$ in reasoning domains on RM benchmarks and best-of-N sampling on GSM8K, with N=16, are presented in Table~\ref{tab:mistral_result}. Our method improves math performance by 30\% on RewardBench and 3\% on Auto-J Eval. Additionally, we enhance reranking performance on GSM8K by 2.4\%. These results demonstrate the adaptability of our methods to different model architectures. The results for different $\lambda$ are presented in Appendix~\ref{sec:appendix:full}.

\begin{table}[t!]
    \centering
    \resizebox{\columnwidth}{!}{
        \begin{tabular}{l c c c c c}
        \toprule
        \multirow{2.5}{*}{\textbf{Model}} & \multicolumn{2}{c}{\textbf{Reward Bench}} & \multicolumn{2}{c}{\textbf{Auto-J Eval}} & \textbf{Best-of-16} \\
        \cmidrule(lr){2-3} \cmidrule(lr){4-5} \cmidrule(lr){6-6} & Code & Math & Code & Math & GSM8K \\
        \midrule
        Mistral RM & \textbf{93.5} & 55.0 & \textbf{88.1} & 87.5 &  44.2 \\
        + MAmmoTH2-Plus & 92.6 & \textbf{85.0} & \textbf{88.1} & \textbf{90.6} & \textbf{46.6} \\
        \bottomrule
        \end{tabular}
    }   
    \caption{Performance of Mistral-based models on various benchmarks and best-of-16 results. Our methods show improvements across RM benchmarks and in best-of-16 sampling on GSM8K.}    \label{tab:mistral_result}
\end{table}

\paragraph{Integrating Multiple Domains}
To evaluate DogeRM's capability of integrating knowledge from multiple domains, we experimented by merging MAmmoTH~\citep{yue2024mammoth} and the Code model into LLaMA-2 RM. We heuristically set the weight factors for MAmmoTH, the Code model, and LLaMA-2 RM at 0.2, 0.2, and 0.6, respectively. The evaluation results, presented in Table~\ref{tab:multi_domain_result}, indicate that merging models from multiple related domains can indeed enhance performance in those domains.

\begin{table}[t!]
    \centering
    \resizebox{\columnwidth}{!}{
        \begin{tabular}{l c c c c c c}
        \toprule
        \multirow{2.5}{*}{\textbf{Model}} & \multicolumn{2}{c}{\textbf{Reward Bench}} & \multicolumn{2}{c}{\textbf{Auto-J Eval}} & \multicolumn{2}{c}{\textbf{Best-of-16}} \\
        \cmidrule(lr){2-3} \cmidrule(lr){4-5} \cmidrule(lr){6-7} 
        & Code & Math & Code & Math & GSM8K & MBPP \\
        \midrule
        LLaMA-2 RM & 78.9 & 68.2 & 76.2 & 84.2 & 35.3 & \textbf{17.2} \\
        + Math \& Code
        & \multicolumn{1}{c}{\textbf{83.0}} 
        & \multicolumn{1}{c}{\textbf{85.2}} 
        & \multicolumn{1}{c}{\textbf{81.0}} 
        & \multicolumn{1}{c}{\textbf{87.5}} 
        & \multicolumn{1}{c}{\textbf{39.5}} 
        & \multicolumn{1}{c}{17.0} \\
        \bottomrule
        \end{tabular}
    }
    \caption{Performance of merging LLaMA-2 RM with MAmmoTH-7B and the Code model on various benchmarks and best-of-16 results. Our methods show improvements across RM benchmarks and in best-of-16 sampling on GSM8K.}
    \label{tab:multi_domain_result}
\end{table}

\section{Conclusion}

In this work, we introduce a novel approach, DogeRM, which integrates domain knowledge into RM by merging it with the domain-specific SFT models. We demonstrate that DogeRM enhances performance on math and coding benchmarks and can be generalized to different model architectures. A series of analyses show that DogeRM effectively affects the reward signal corresponding to chosen and rejected prompts. The results highlight DogeRM’s potential to enhance model alignment and generation verification through model merging, offering promising results across various benchmarks.

\section*{Limitations}

There are several limitations in our work: (1) Our framework has been tested exclusively in the math and coding domains, leaving other areas such as medicine, finance, and law unexplored. In Section~\ref{subsec:analysis}, we demonstrated the effectiveness of merging domain-specific models for math and coding into RM. However, the integration of models from multiple orthogonal domains remains an area for future investigation. (2) Our method was tested exclusively on 7B models, and we have not evaluated its performance on models of larger or smaller sizes. (3) While our framework is compatible with various merging techniques, such as TIES-Merge~\citep{yadav2024ties}, we have not thoroughly examined the impact of these more advanced methods. In this work, we focused on demonstrating the core idea—improving RM performance in a target domain by merging with a domain-specific language model. To keep our approach straightforward, we used weighted averaging, which, in the case of two models, can be understood as a form of task arithmetic. Despite the simplicity of this method, it has already led to notable performance gains. However, the effectiveness of more sophisticated merging techniques remains unexplored, and we leave this investigation for future work. (4) The models being merged must share the same architecture, a limitation common to most model merging algorithms~\citep{wortsman2022model,ilharco2023editing,yadav2024ties}. Recently, evolutionary model merging~\citep{akiba2024evolutionary} has been proposed as a solution for merging models with different architectures. Investigating the merging of models with varying architectures remains a topic for future research. (5) Due to the sensitivity of RLHF to hyperparameter choices and our limited computational resources, we did not implement RLHF algorithms such as PPO~\citep{schulman2017proximal} or RLOO~\citep{ahmadian-etal-2024-back} in this work. Exploring the integration of DogeRM within RLHF frameworks is left for future work.

\section*{Ethics Statement}

While our method effectively equips reward models with domain knowledge, it does not eliminate the inherent biases within these models. Further investigation is needed to explore the impact of these inherited biases in the original reward models.

\section*{Acknowledgments}
We thank the reviewers for their insightful comments.
This work was financially supported by the National Science and Technology Council (NSTC) in Taiwan, under Grants 111-2222-E-002-013-MY3 and 112-2223-E002-012-MY5. 
We thank to National Center for High-performance Computing (NCHC) of National Applied Research Laboratories (NARLabs) in Taiwan for providing computational and storage resources.
 We are also grateful to Yen-Ting Lin, Wei-Lin Chen, Chao-Wei Huang and Wan-Xuan Zhou from National Taiwan University for their insightful discussions and valuable advice on Figure~\ref{fig:overview}.

\bibliography{custom}

\begin{thebibliography}{72}
\providecommand{\natexlab}[1]{#1}

\bibitem[{Achiam et~al.(2023)Achiam, Adler, Agarwal, Ahmad, Akkaya, Aleman, Almeida, Altenschmidt, Altman, Anadkat et~al.}]{achiam2023gpt}
Josh Achiam, Steven Adler, Sandhini Agarwal, Lama Ahmad, Ilge Akkaya, Florencia~Leoni Aleman, Diogo Almeida, Janko Altenschmidt, Sam Altman, Shyamal Anadkat, et~al. 2023.
\newblock Gpt-4 technical report.
\newblock \emph{arXiv preprint arXiv:2303.08774}.

\bibitem[{Ahmadian et~al.(2024)Ahmadian, Cremer, Gall{\'e}, Fadaee, Kreutzer, Pietquin, {\"U}st{\"u}n, and Hooker}]{ahmadian-etal-2024-back}
Arash Ahmadian, Chris Cremer, Matthias Gall{\'e}, Marzieh Fadaee, Julia Kreutzer, Olivier Pietquin, Ahmet {\"U}st{\"u}n, and Sara Hooker. 2024.
\newblock \href {https://doi.org/10.18653/v1/2024.acl-long.662} {Back to basics: Revisiting {REINFORCE}-style optimization for learning from human feedback in {LLM}s}.
\newblock In \emph{Proceedings of the 62nd Annual Meeting of the Association for Computational Linguistics (Volume 1: Long Papers)}, pages 12248--12267, Bangkok, Thailand. Association for Computational Linguistics.

\bibitem[{Akiba et~al.(2024)Akiba, Shing, Tang, Sun, and Ha}]{akiba2024evolutionary}
Takuya Akiba, Makoto Shing, Yujin Tang, Qi~Sun, and David Ha. 2024.
\newblock \href {https://arxiv.org/abs/2403.13187} {Evolutionary optimization of model merging recipes}.
\newblock \emph{Preprint}, arXiv:2403.13187.

\bibitem[{Askell et~al.(2021)Askell, Bai, Chen, Drain, Ganguli, Henighan, Jones, Joseph, Mann, DasSarma, Elhage, Hatfield-Dodds, Hernandez, Kernion, Ndousse, Olsson, Amodei, Brown, Clark, McCandlish, Olah, and Kaplan}]{askell2021general}
Amanda Askell, Yuntao Bai, Anna Chen, Dawn Drain, Deep Ganguli, Tom Henighan, Andy Jones, Nicholas Joseph, Ben Mann, Nova DasSarma, Nelson Elhage, Zac Hatfield-Dodds, Danny Hernandez, Jackson Kernion, Kamal Ndousse, Catherine Olsson, Dario Amodei, Tom Brown, Jack Clark, Sam McCandlish, Chris Olah, and Jared Kaplan. 2021.
\newblock \href {https://arxiv.org/abs/2112.00861} {A general language assistant as a laboratory for alignment}.
\newblock \emph{Preprint}, arXiv:2112.00861.

\bibitem[{Austin et~al.(2021)Austin, Odena, Nye, Bosma, Michalewski, Dohan, Jiang, Cai, Terry, Le et~al.}]{austin2021program}
Jacob Austin, Augustus Odena, Maxwell Nye, Maarten Bosma, Henryk Michalewski, David Dohan, Ellen Jiang, Carrie Cai, Michael Terry, Quoc Le, et~al. 2021.
\newblock Program synthesis with large language models.
\newblock \emph{arXiv preprint arXiv:2108.07732}.

\bibitem[{Bai et~al.(2022)Bai, Jones, Ndousse, Askell, Chen, DasSarma, Drain, Fort, Ganguli, Henighan, Joseph, Kadavath, Kernion, Conerly, El-Showk, Elhage, Hatfield-Dodds, Hernandez, Hume, Johnston, Kravec, Lovitt, Nanda, Olsson, Amodei, Brown, Clark, McCandlish, Olah, Mann, and Kaplan}]{bai2022training}
Yuntao Bai, Andy Jones, Kamal Ndousse, Amanda Askell, Anna Chen, Nova DasSarma, Dawn Drain, Stanislav Fort, Deep Ganguli, Tom Henighan, Nicholas Joseph, Saurav Kadavath, Jackson Kernion, Tom Conerly, Sheer El-Showk, Nelson Elhage, Zac Hatfield-Dodds, Danny Hernandez, Tristan Hume, Scott Johnston, Shauna Kravec, Liane Lovitt, Neel Nanda, Catherine Olsson, Dario Amodei, Tom Brown, Jack Clark, Sam McCandlish, Chris Olah, Ben Mann, and Jared Kaplan. 2022.
\newblock \href {https://arxiv.org/abs/2204.05862} {Training a helpful and harmless assistant with reinforcement learning from human feedback}.
\newblock \emph{Preprint}, arXiv:2204.05862.

\bibitem[{Bartolome et~al.(2023)Bartolome, Martin, and Vila}]{notus2023}
Alvaro Bartolome, Gabriel Martin, and Daniel Vila. 2023.
\newblock Notus.
\newblock \url{https://github.com/argilla-io/notus}.

\bibitem[{Beeching et~al.(2023)Beeching, Fourrier, Habib, Han, Lambert, Rajani, Sanseviero, Tunstall, and Wolf}]{open-llm-leaderboard}
Edward Beeching, Clémentine Fourrier, Nathan Habib, Sheon Han, Nathan Lambert, Nazneen Rajani, Omar Sanseviero, Lewis Tunstall, and Thomas Wolf. 2023.
\newblock Open llm leaderboard.
\newblock \url{https://huggingface.co/spaces/open-llm-leaderboard/open_llm_leaderboard}.

\bibitem[{Chen et~al.(2021)Chen, Tworek, Jun, Yuan, Pinto, Kaplan, Edwards, Burda, Joseph, Brockman et~al.}]{chen2021codex}
Mark Chen, Jerry Tworek, Heewoo Jun, Qiming Yuan, Henrique Ponde de~Oliveira Pinto, Jared Kaplan, Harri Edwards, Yuri Burda, Nicholas Joseph, Greg Brockman, et~al. 2021.
\newblock Evaluating large language models trained on code.
\newblock \emph{arXiv preprint arXiv:2107.03374}.

\bibitem[{Chiang et~al.(2023)Chiang, Li, Lin, Sheng, Wu, Zhang, Zheng, Zhuang, Zhuang, Gonzalez, Stoica, and Xing}]{vicuna2023}
Wei-Lin Chiang, Zhuohan Li, Zi~Lin, Ying Sheng, Zhanghao Wu, Hao Zhang, Lianmin Zheng, Siyuan Zhuang, Yonghao Zhuang, Joseph~E. Gonzalez, Ion Stoica, and Eric~P. Xing. 2023.
\newblock \href {https://lmsys.org/blog/2023-03-30-vicuna/} {Vicuna: An open-source chatbot impressing gpt-4 with 90\%* chatgpt quality}.

\bibitem[{Christiano et~al.(2017)Christiano, Leike, Brown, Martic, Legg, and Amodei}]{christiano2017deep}
Paul~F Christiano, Jan Leike, Tom Brown, Miljan Martic, Shane Legg, and Dario Amodei. 2017.
\newblock Deep reinforcement learning from human preferences.
\newblock \emph{Advances in neural information processing systems}, 30.

\bibitem[{Cobbe et~al.(2021)Cobbe, Kosaraju, Bavarian, Chen, Jun, Kaiser, Plappert, Tworek, Hilton, Nakano et~al.}]{cobbe2021training}
Karl Cobbe, Vineet Kosaraju, Mohammad Bavarian, Mark Chen, Heewoo Jun, Lukasz Kaiser, Matthias Plappert, Jerry Tworek, Jacob Hilton, Reiichiro Nakano, et~al. 2021.
\newblock Training verifiers to solve math word problems.
\newblock \emph{arXiv preprint arXiv:2110.14168}.

\bibitem[{Cui et~al.(2024)Cui, Yuan, Ding, Yao, He, Zhu, Ni, Xie, Xie, Lin et~al.}]{cui2024ultrafeedback}
Ganqu Cui, Lifan Yuan, Ning Ding, Guanming Yao, Bingxiang He, Wei Zhu, Yuan Ni, Guotong Xie, Ruobing Xie, Yankai Lin, et~al. 2024.
\newblock Ultrafeedback: Boosting language models with scaled ai feedback.
\newblock In \emph{Forty-first International Conference on Machine Learning}.

\bibitem[{Ding et~al.(2023)Ding, Chen, Xu, Qin, Hu, Liu, Sun, and Zhou}]{ding2023enhancing}
Ning Ding, Yulin Chen, Bokai Xu, Yujia Qin, Shengding Hu, Zhiyuan Liu, Maosong Sun, and Bowen Zhou. 2023.
\newblock \href {https://openreview.net/forum?id=oEsYs3WRc3} {Enhancing chat language models by scaling high-quality instructional conversations}.
\newblock In \emph{The 2023 Conference on Empirical Methods in Natural Language Processing}.

\bibitem[{Dubois et~al.(2023)Dubois, Li, Taori, Zhang, Gulrajani, Ba, Guestrin, Liang, and Hashimoto}]{dubois2023alpacafarm}
Yann Dubois, Chen~Xuechen Li, Rohan Taori, Tianyi Zhang, Ishaan Gulrajani, Jimmy Ba, Carlos Guestrin, Percy~S Liang, and Tatsunori~B Hashimoto. 2023.
\newblock Alpacafarm: A simulation framework for methods that learn from human feedback.
\newblock \emph{Advances in Neural Information Processing Systems}, 36.

\bibitem[{Ethayarajh et~al.(2022)Ethayarajh, Choi, and Swayamdipta}]{pmlr-v162-ethayarajh22a}
Kawin Ethayarajh, Yejin Choi, and Swabha Swayamdipta. 2022.
\newblock \href {https://proceedings.mlr.press/v162/ethayarajh22a.html} {Understanding dataset difficulty with $\mathcal{V}$-usable information}.
\newblock In \emph{Proceedings of the 39th International Conference on Machine Learning}, volume 162 of \emph{Proceedings of Machine Learning Research}, pages 5988--6008. PMLR.

\bibitem[{Gao et~al.(2023)Gao, Tow, Abbasi, Biderman, Black, DiPofi, Foster, Golding, Hsu, Le~Noac'h, Li, McDonell, Muennighoff, Ociepa, Phang, Reynolds, Schoelkopf, Skowron, Sutawika, Tang, Thite, Wang, Wang, and Zou}]{eval-harness}
Leo Gao, Jonathan Tow, Baber Abbasi, Stella Biderman, Sid Black, Anthony DiPofi, Charles Foster, Laurence Golding, Jeffrey Hsu, Alain Le~Noac'h, Haonan Li, Kyle McDonell, Niklas Muennighoff, Chris Ociepa, Jason Phang, Laria Reynolds, Hailey Schoelkopf, Aviya Skowron, Lintang Sutawika, Eric Tang, Anish Thite, Ben Wang, Kevin Wang, and Andy Zou. 2023.
\newblock \href {https://doi.org/10.5281/zenodo.10256836} {A framework for few-shot language model evaluation}.

\bibitem[{Havrilla(2023)}]{syntheticgptj}
Alex Havrilla. 2023.
\newblock synthetic-instruct-gptj-pairwise.
\newblock \url{https://huggingface.co/datasets/Dahoas/synthetic-instruct-gptj-pairwise}.

\bibitem[{Hendrycks et~al.(2021)Hendrycks, Burns, Kadavath, Arora, Basart, Tang, Song, and Steinhardt}]{hendrycks2021measuring}
Dan Hendrycks, Collin Burns, Saurav Kadavath, Akul Arora, Steven Basart, Eric Tang, Dawn Song, and Jacob Steinhardt. 2021.
\newblock \href {https://openreview.net/forum?id=7Bywt2mQsCe} {Measuring mathematical problem solving with the {MATH} dataset}.
\newblock In \emph{Thirty-fifth Conference on Neural Information Processing Systems Datasets and Benchmarks Track (Round 2)}.

\bibitem[{Hu et~al.(2023)Hu, Luo, Wang, Cheng, Liu, and Sun}]{hu-etal-2023-wont}
Shengding Hu, Yifan Luo, Huadong Wang, Xingyi Cheng, Zhiyuan Liu, and Maosong Sun. 2023.
\newblock \href {https://doi.org/10.18653/v1/2023.acl-long.309} {Won{'}t get fooled again: Answering questions with false premises}.
\newblock In \emph{Proceedings of the 61st Annual Meeting of the Association for Computational Linguistics (Volume 1: Long Papers)}, pages 5626--5643, Toronto, Canada. Association for Computational Linguistics.

\bibitem[{Huang et~al.(2024)Huang, Li, Hsu, Chen, Lin, Hsiao, Tsai, and Lee}]{huang-etal-2024-chat}
Shih-Cheng Huang, Pin-Zu Li, Yu-chi Hsu, Kuang-Ming Chen, Yu~Tung Lin, Shih-Kai Hsiao, Richard Tsai, and Hung-yi Lee. 2024.
\newblock \href {https://doi.org/10.18653/v1/2024.acl-long.590} {Chat vector: A simple approach to equip {LLM}s with instruction following and model alignment in new languages}.
\newblock In \emph{Proceedings of the 62nd Annual Meeting of the Association for Computational Linguistics (Volume 1: Long Papers)}, pages 10943--10959, Bangkok, Thailand. Association for Computational Linguistics.

\bibitem[{Ilharco et~al.(2023)Ilharco, Ribeiro, Wortsman, Schmidt, Hajishirzi, and Farhadi}]{ilharco2023editing}
Gabriel Ilharco, Marco~Tulio Ribeiro, Mitchell Wortsman, Ludwig Schmidt, Hannaneh Hajishirzi, and Ali Farhadi. 2023.
\newblock \href {https://openreview.net/forum?id=6t0Kwf8-jrj} {Editing models with task arithmetic}.
\newblock In \emph{The Eleventh International Conference on Learning Representations}.

\bibitem[{Jang et~al.(2023)Jang, Kim, Lin, Wang, Hessel, Zettlemoyer, Hajishirzi, Choi, and Ammanabrolu}]{jang2023personalized}
Joel Jang, Seungone Kim, Bill~Yuchen Lin, Yizhong Wang, Jack Hessel, Luke Zettlemoyer, Hannaneh Hajishirzi, Yejin Choi, and Prithviraj Ammanabrolu. 2023.
\newblock Personalized soups: Personalized large language model alignment via post-hoc parameter merging.
\newblock \emph{arXiv preprint arXiv:2310.11564}.

\bibitem[{Ji et~al.(2023)Ji, Liu, Dai, Pan, Zhang, Bian, Chen, Sun, Wang, and Yang}]{pkualignment}
Jiaming Ji, Mickel Liu, Josef Dai, Xuehai Pan, Chi Zhang, Ce~Bian, Boyuan Chen, Ruiyang Sun, Yizhou Wang, and Yaodong Yang. 2023.
\newblock \href {https://proceedings.neurips.cc/paper_files/paper/2023/file/4dbb61cb68671edc4ca3712d70083b9f-Paper-Datasets_and_Benchmarks.pdf} {Beavertails: Towards improved safety alignment of llm via a human-preference dataset}.
\newblock In \emph{Advances in Neural Information Processing Systems}, volume~36, pages 24678--24704. Curran Associates, Inc.

\bibitem[{Jiang et~al.(2023)Jiang, Sablayrolles, Mensch, Bamford, Chaplot, de~las Casas, Bressand, Lengyel, Lample, Saulnier, Lavaud, Lachaux, Stock, Scao, Lavril, Wang, Lacroix, and Sayed}]{jiang2023mistral}
Albert~Q. Jiang, Alexandre Sablayrolles, Arthur Mensch, Chris Bamford, Devendra~Singh Chaplot, Diego de~las Casas, Florian Bressand, Gianna Lengyel, Guillaume Lample, Lucile Saulnier, Lélio~Renard Lavaud, Marie-Anne Lachaux, Pierre Stock, Teven~Le Scao, Thibaut Lavril, Thomas Wang, Timothée Lacroix, and William~El Sayed. 2023.
\newblock \href {https://arxiv.org/abs/2310.06825} {Mistral 7b}.
\newblock \emph{Preprint}, arXiv:2310.06825.

\bibitem[{Jin et~al.(2023)Jin, Ren, Preotiuc-Pietro, and Cheng}]{jin2023dataless}
Xisen Jin, Xiang Ren, Daniel Preotiuc-Pietro, and Pengxiang Cheng. 2023.
\newblock \href {https://openreview.net/forum?id=FCnohuR6AnM} {Dataless knowledge fusion by merging weights of language models}.
\newblock In \emph{The Eleventh International Conference on Learning Representations}.

\bibitem[{Kwon et~al.(2023)Kwon, Li, Zhuang, Sheng, Zheng, Yu, Gonzalez, Zhang, and Stoica}]{kwon2023efficient}
Woosuk Kwon, Zhuohan Li, Siyuan Zhuang, Ying Sheng, Lianmin Zheng, Cody~Hao Yu, Joseph~E. Gonzalez, Hao Zhang, and Ion Stoica. 2023.
\newblock Efficient memory management for large language model serving with pagedattention.
\newblock In \emph{Proceedings of the ACM SIGOPS 29th Symposium on Operating Systems Principles}.

\bibitem[{Lambert et~al.(2024)Lambert, Pyatkin, Morrison, Miranda, Lin, Chandu, Dziri, Kumar, Zick, Choi et~al.}]{lambert2024rewardbench}
Nathan Lambert, Valentina Pyatkin, Jacob Morrison, LJ~Miranda, Bill~Yuchen Lin, Khyathi Chandu, Nouha Dziri, Sachin Kumar, Tom Zick, Yejin Choi, et~al. 2024.
\newblock Rewardbench: Evaluating reward models for language modeling.
\newblock \emph{arXiv preprint arXiv:2403.13787}.

\bibitem[{Li et~al.(2024)Li, Sun, Yuan, Fan, hai zhao, and Liu}]{li2024generative}
Junlong Li, Shichao Sun, Weizhe Yuan, Run-Ze Fan, hai zhao, and Pengfei Liu. 2024.
\newblock \href {https://openreview.net/forum?id=gtkFw6sZGS} {Generative judge for evaluating alignment}.
\newblock In \emph{The Twelfth International Conference on Learning Representations}.

\bibitem[{Li et~al.(2023)Li, Zhang, Dubois, Taori, Gulrajani, Guestrin, Liang, and Hashimoto}]{alpaca_eval}
Xuechen Li, Tianyi Zhang, Yann Dubois, Rohan Taori, Ishaan Gulrajani, Carlos Guestrin, Percy Liang, and Tatsunori~B. Hashimoto. 2023.
\newblock Alpacaeval: An automatic evaluator of instruction-following models.
\newblock \url{https://github.com/tatsu-lab/alpaca_eval}.

\bibitem[{Lightman et~al.(2024)Lightman, Kosaraju, Burda, Edwards, Baker, Lee, Leike, Schulman, Sutskever, and Cobbe}]{lightman2024lets}
Hunter Lightman, Vineet Kosaraju, Yuri Burda, Harrison Edwards, Bowen Baker, Teddy Lee, Jan Leike, John Schulman, Ilya Sutskever, and Karl Cobbe. 2024.
\newblock \href {https://openreview.net/forum?id=v8L0pN6EOi} {Let's verify step by step}.
\newblock In \emph{The Twelfth International Conference on Learning Representations}.

\bibitem[{Lin et~al.(2022)Lin, Hilton, and Evans}]{lin-etal-2022-truthfulqa}
Stephanie Lin, Jacob Hilton, and Owain Evans. 2022.
\newblock \href {https://doi.org/10.18653/v1/2022.acl-long.229} {{T}ruthful{QA}: Measuring how models mimic human falsehoods}.
\newblock In \emph{Proceedings of the 60th Annual Meeting of the Association for Computational Linguistics (Volume 1: Long Papers)}, pages 3214--3252, Dublin, Ireland. Association for Computational Linguistics.

\bibitem[{Longpre et~al.(2023)Longpre, Hou, Vu, Webson, Chung, Tay, Zhou, Le, Zoph, Wei et~al.}]{longpre2023flan}
Shayne Longpre, Le~Hou, Tu~Vu, Albert Webson, Hyung~Won Chung, Yi~Tay, Denny Zhou, Quoc~V Le, Barret Zoph, Jason Wei, et~al. 2023.
\newblock The flan collection: Designing data and methods for effective instruction tuning.
\newblock In \emph{International Conference on Machine Learning}, pages 22631--22648. PMLR.

\bibitem[{Luo et~al.(2024)Luo, Xu, Zhao, Sun, Geng, Hu, Tao, Ma, Lin, and Jiang}]{luo2023wizardcoder}
Ziyang Luo, Can Xu, Pu~Zhao, Qingfeng Sun, Xiubo Geng, Wenxiang Hu, Chongyang Tao, Jing Ma, Qingwei Lin, and Daxin Jiang. 2024.
\newblock \href {https://openreview.net/forum?id=UnUwSIgK5W} {Wizardcoder: Empowering code large language models with evol-instruct}.
\newblock In \emph{The Twelfth International Conference on Learning Representations}.

\bibitem[{Matena and Raffel(2022)}]{matena2022merging}
Michael~S Matena and Colin~A Raffel. 2022.
\newblock Merging models with fisher-weighted averaging.
\newblock \emph{Advances in Neural Information Processing Systems}, 35:17703--17716.

\bibitem[{Menick et~al.(2022)Menick, Trebacz, Mikulik, Aslanides, Song, Chadwick, Glaese, Young, Campbell-Gillingham, Irving, and McAleese}]{menick2022teaching}
Jacob Menick, Maja Trebacz, Vladimir Mikulik, John Aslanides, Francis Song, Martin Chadwick, Mia Glaese, Susannah Young, Lucy Campbell-Gillingham, Geoffrey Irving, and Nat McAleese. 2022.
\newblock \href {https://arxiv.org/abs/2203.11147} {Teaching language models to support answers with verified quotes}.
\newblock \emph{Preprint}, arXiv:2203.11147.

\bibitem[{Muennighoff et~al.(2024)Muennighoff, Liu, Zebaze, Zheng, Hui, Zhuo, Singh, Tang, Werra, and Longpre}]{muennighoff2024octopack}
Niklas Muennighoff, Qian Liu, Armel~Randy Zebaze, Qinkai Zheng, Binyuan Hui, Terry~Yue Zhuo, Swayam Singh, Xiangru Tang, Leandro~Von Werra, and Shayne Longpre. 2024.
\newblock \href {https://openreview.net/forum?id=mw1PWNSWZP} {Octopack: Instruction tuning code large language models}.
\newblock In \emph{The Twelfth International Conference on Learning Representations}.

\bibitem[{Nakano et~al.(2022)Nakano, Hilton, Balaji, Wu, Ouyang, Kim, Hesse, Jain, Kosaraju, Saunders, Jiang, Cobbe, Eloundou, Krueger, Button, Knight, Chess, and Schulman}]{nakano2022webgpt}
Reiichiro Nakano, Jacob Hilton, Suchir Balaji, Jeff Wu, Long Ouyang, Christina Kim, Christopher Hesse, Shantanu Jain, Vineet Kosaraju, William Saunders, Xu~Jiang, Karl Cobbe, Tyna Eloundou, Gretchen Krueger, Kevin Button, Matthew Knight, Benjamin Chess, and John Schulman. 2022.
\newblock \href {https://arxiv.org/abs/2112.09332} {Webgpt: Browser-assisted question-answering with human feedback}.
\newblock \emph{Preprint}, arXiv:2112.09332.

\bibitem[{OpenAI(2023)}]{openai2023textdavinci003}
OpenAI. 2023.
\newblock Openai gpt-3 api \texttt{text-davinci-003} (deprecated).
\newblock Accessed: 2023-06-15. Available at: \url{https://beta.openai.com/docs/models/gpt-3}.

\bibitem[{Ouyang et~al.(2022)Ouyang, Wu, Jiang, Almeida, Wainwright, Mishkin, Zhang, Agarwal, Slama, Ray, Schulman, Hilton, Kelton, Miller, Simens, Askell, Welinder, Christiano, Leike, and Lowe}]{instructgpt}
Long Ouyang, Jeffrey Wu, Xu~Jiang, Diogo Almeida, Carroll Wainwright, Pamela Mishkin, Chong Zhang, Sandhini Agarwal, Katarina Slama, Alex Ray, John Schulman, Jacob Hilton, Fraser Kelton, Luke Miller, Maddie Simens, Amanda Askell, Peter Welinder, Paul~F Christiano, Jan Leike, and Ryan Lowe. 2022.
\newblock \href {https://proceedings.neurips.cc/paper_files/paper/2022/file/b1efde53be364a73914f58805a001731-Paper-Conference.pdf} {Training language models to follow instructions with human feedback}.
\newblock In \emph{Advances in Neural Information Processing Systems}, volume~35, pages 27730--27744. Curran Associates, Inc.

\bibitem[{Ramamurthy et~al.(2023)Ramamurthy, Ammanabrolu, Brantley, Hessel, Sifa, Bauckhage, Hajishirzi, and Choi}]{ramamurthy2023is}
Rajkumar Ramamurthy, Prithviraj Ammanabrolu, Kiant{\'e} Brantley, Jack Hessel, Rafet Sifa, Christian Bauckhage, Hannaneh Hajishirzi, and Yejin Choi. 2023.
\newblock \href {https://openreview.net/forum?id=8aHzds2uUyB} {Is reinforcement learning (not) for natural language processing: Benchmarks, baselines, and building blocks for natural language policy optimization}.
\newblock In \emph{The Eleventh International Conference on Learning Representations}.

\bibitem[{Rame et~al.(2024{\natexlab{a}})Rame, Couairon, Dancette, Gaya, Shukor, Soulier, and Cord}]{rame2024rewarded}
Alexandre Rame, Guillaume Couairon, Corentin Dancette, Jean-Baptiste Gaya, Mustafa Shukor, Laure Soulier, and Matthieu Cord. 2024{\natexlab{a}}.
\newblock Rewarded soups: towards pareto-optimal alignment by interpolating weights fine-tuned on diverse rewards.
\newblock \emph{Advances in Neural Information Processing Systems}, 36.

\bibitem[{Rame et~al.(2024{\natexlab{b}})Rame, Vieillard, Hussenot, Dadashi, Cideron, Bachem, and Ferret}]{ramé2024warm}
Alexandre Rame, Nino Vieillard, Leonard Hussenot, Robert Dadashi, Geoffrey Cideron, Olivier Bachem, and Johan Ferret. 2024{\natexlab{b}}.
\newblock \href {https://openreview.net/forum?id=s7RDnNUJy6} {{WARM}: On the benefits of weight averaged reward models}.
\newblock In \emph{Forty-first International Conference on Machine Learning}.

\bibitem[{Ray2333(2024)}]{mistralrm}
Ray2333. 2024.
\newblock reward-model-mistral-7b-instruct-unified-feedback.
\newblock \href{https://huggingface.co/Ray2333/reward-model-Mistral-7B-instruct-Unified-Feedback}{Ray2333/reward-model-Mistral-7B-instruct-Unified-Feedback}.
\newblock \raggedright Accessed: June 2024.

\bibitem[{R{\"o}ttger et~al.(2024)R{\"o}ttger, Kirk, Vidgen, Attanasio, Bianchi, and Hovy}]{rottger-etal-2024-xstest}
Paul R{\"o}ttger, Hannah Kirk, Bertie Vidgen, Giuseppe Attanasio, Federico Bianchi, and Dirk Hovy. 2024.
\newblock \href {https://doi.org/10.18653/v1/2024.naacl-long.301} {{XST}est: A test suite for identifying exaggerated safety behaviours in large language models}.
\newblock In \emph{Proceedings of the 2024 Conference of the North American Chapter of the Association for Computational Linguistics: Human Language Technologies (Volume 1: Long Papers)}, pages 5377--5400, Mexico City, Mexico. Association for Computational Linguistics.

\bibitem[{Schulman et~al.(2017)Schulman, Wolski, Dhariwal, Radford, and Klimov}]{schulman2017proximal}
John Schulman, Filip Wolski, Prafulla Dhariwal, Alec Radford, and Oleg Klimov. 2017.
\newblock Proximal policy optimization algorithms.
\newblock \emph{arXiv preprint arXiv:1707.06347}.

\bibitem[{Stiennon et~al.(2020)Stiennon, Ouyang, Wu, Ziegler, Lowe, Voss, Radford, Amodei, and Christiano}]{openaisummarize}
Nisan Stiennon, Long Ouyang, Jeffrey Wu, Daniel Ziegler, Ryan Lowe, Chelsea Voss, Alec Radford, Dario Amodei, and Paul~F Christiano. 2020.
\newblock \href {https://proceedings.neurips.cc/paper_files/paper/2020/file/1f89885d556929e98d3ef9b86448f951-Paper.pdf} {Learning to summarize with human feedback}.
\newblock In \emph{Advances in Neural Information Processing Systems}, volume~33, pages 3008--3021. Curran Associates, Inc.

\bibitem[{Taori et~al.(2023)Taori, Gulrajani, Zhang, Dubois, Li, Guestrin, Liang, and Hashimoto}]{alpaca}
Rohan Taori, Ishaan Gulrajani, Tianyi Zhang, Yann Dubois, Xuechen Li, Carlos Guestrin, Percy Liang, and Tatsunori~B. Hashimoto. 2023.
\newblock Stanford alpaca: An instruction-following llama model.
\newblock \url{https://github.com/tatsu-lab/stanford_alpaca}.

\bibitem[{Team et~al.(2023)Team, Anil, Borgeaud, Wu, Alayrac, Yu, Soricut, Schalkwyk, Dai, Hauth et~al.}]{team2023gemini}
Gemini Team, Rohan Anil, Sebastian Borgeaud, Yonghui Wu, Jean-Baptiste Alayrac, Jiahui Yu, Radu Soricut, Johan Schalkwyk, Andrew~M Dai, Anja Hauth, et~al. 2023.
\newblock Gemini: a family of highly capable multimodal models.
\newblock \emph{arXiv preprint arXiv:2312.11805}.

\bibitem[{theblackcat102(2023)}]{evolcode}
theblackcat102. 2023.
\newblock The evolved code alpaca dataset.
\newblock \url{https://huggingface.co/datasets/theblackcat102/evol-codealpaca-v1}.

\bibitem[{Touvron et~al.(2023)Touvron, Martin, Stone, Albert, Almahairi, Babaei, Bashlykov, Batra, Bhargava, Bhosale et~al.}]{touvron2023llama}
Hugo Touvron, Louis Martin, Kevin Stone, Peter Albert, Amjad Almahairi, Yasmine Babaei, Nikolay Bashlykov, Soumya Batra, Prajjwal Bhargava, Shruti Bhosale, et~al. 2023.
\newblock Llama 2: Open foundation and fine-tuned chat models.
\newblock \emph{arXiv preprint arXiv:2307.09288}.

\bibitem[{von Werra et~al.(2020)von Werra, Belkada, Tunstall, Beeching, Thrush, Lambert, and Huang}]{vonwerra2022trl}
Leandro von Werra, Younes Belkada, Lewis Tunstall, Edward Beeching, Tristan Thrush, Nathan Lambert, and Shengyi Huang. 2020.
\newblock Trl: Transformer reinforcement learning.
\newblock \url{https://github.com/huggingface/trl}.

\bibitem[{Wang et~al.(2024{\natexlab{a}})Wang, Lin, Xiong, Yang, Diao, Qiu, Zhao, and Zhang}]{wang-etal-2024-arithmetic}
Haoxiang Wang, Yong Lin, Wei Xiong, Rui Yang, Shizhe Diao, Shuang Qiu, Han Zhao, and Tong Zhang. 2024{\natexlab{a}}.
\newblock \href {https://doi.org/10.18653/v1/2024.acl-long.468} {Arithmetic control of {LLM}s for diverse user preferences: Directional preference alignment with multi-objective rewards}.
\newblock In \emph{Proceedings of the 62nd Annual Meeting of the Association for Computational Linguistics (Volume 1: Long Papers)}, pages 8642--8655, Bangkok, Thailand. Association for Computational Linguistics.

\bibitem[{Wang et~al.(2024{\natexlab{b}})Wang, Li, Han, Nakov, and Baldwin}]{wang-etal-2024-answer}
Yuxia Wang, Haonan Li, Xudong Han, Preslav Nakov, and Timothy Baldwin. 2024{\natexlab{b}}.
\newblock \href {https://aclanthology.org/2024.findings-eacl.61} {Do-not-answer: Evaluating safeguards in {LLM}s}.
\newblock In \emph{Findings of the Association for Computational Linguistics: EACL 2024}, pages 896--911, St. Julian{'}s, Malta. Association for Computational Linguistics.

\bibitem[{Wang et~al.(2024{\natexlab{c}})Wang, Dong, Zeng, Adams, Sreedhar, Egert, Delalleau, Scowcroft, Kant, Swope, and Kuchaiev}]{wang-etal-2024-helpsteer}
Zhilin Wang, Yi~Dong, Jiaqi Zeng, Virginia Adams, Makesh~Narsimhan Sreedhar, Daniel Egert, Olivier Delalleau, Jane Scowcroft, Neel Kant, Aidan Swope, and Oleksii Kuchaiev. 2024{\natexlab{c}}.
\newblock \href {https://doi.org/10.18653/v1/2024.naacl-long.185} {{H}elp{S}teer: Multi-attribute helpfulness dataset for {S}teer{LM}}.
\newblock In \emph{Proceedings of the 2024 Conference of the North American Chapter of the Association for Computational Linguistics: Human Language Technologies (Volume 1: Long Papers)}, pages 3371--3384, Mexico City, Mexico. Association for Computational Linguistics.

\bibitem[{Wei et~al.(2024)Wei, Wang, Liu, Ding, and ZHANG}]{wei2024magicoder}
Yuxiang Wei, Zhe Wang, Jiawei Liu, Yifeng Ding, and LINGMING ZHANG. 2024.
\newblock \href {https://openreview.net/forum?id=XUeoOBid3x} {Magicoder: Empowering code generation with {OSS}-instruct}.
\newblock In \emph{Forty-first International Conference on Machine Learning}.

\bibitem[{Wolf et~al.(2020)Wolf, Debut, Sanh, Chaumond, Delangue, Moi, Cistac, Rault, Louf, Funtowicz, Davison, Shleifer, von Platen, Ma, Jernite, Plu, Xu, Le~Scao, Gugger, Drame, Lhoest, and Rush}]{wolf-etal-2020-transformers}
Thomas Wolf, Lysandre Debut, Victor Sanh, Julien Chaumond, Clement Delangue, Anthony Moi, Pierric Cistac, Tim Rault, Remi Louf, Morgan Funtowicz, Joe Davison, Sam Shleifer, Patrick von Platen, Clara Ma, Yacine Jernite, Julien Plu, Canwen Xu, Teven Le~Scao, Sylvain Gugger, Mariama Drame, Quentin Lhoest, and Alexander Rush. 2020.
\newblock \href {https://doi.org/10.18653/v1/2020.emnlp-demos.6} {Transformers: State-of-the-art natural language processing}.
\newblock In \emph{Proceedings of the 2020 Conference on Empirical Methods in Natural Language Processing: System Demonstrations}, pages 38--45, Online. Association for Computational Linguistics.

\bibitem[{Wortsman et~al.(2022)Wortsman, Ilharco, Gadre, Roelofs, Gontijo-Lopes, Morcos, Namkoong, Farhadi, Carmon, Kornblith et~al.}]{wortsman2022model}
Mitchell Wortsman, Gabriel Ilharco, Samir~Ya Gadre, Rebecca Roelofs, Raphael Gontijo-Lopes, Ari~S Morcos, Hongseok Namkoong, Ali Farhadi, Yair Carmon, Simon Kornblith, et~al. 2022.
\newblock Model soups: averaging weights of multiple fine-tuned models improves accuracy without increasing inference time.
\newblock In \emph{International conference on machine learning}, pages 23965--23998. PMLR.

\bibitem[{Wu et~al.(2021)Wu, Ouyang, Ziegler, Stiennon, Lowe, Leike, and Christiano}]{wu2021recursively}
Jeff Wu, Long Ouyang, Daniel~M. Ziegler, Nisan Stiennon, Ryan Lowe, Jan Leike, and Paul Christiano. 2021.
\newblock \href {https://arxiv.org/abs/2109.10862} {Recursively summarizing books with human feedback}.
\newblock \emph{Preprint}, arXiv:2109.10862.

\bibitem[{Wu et~al.(2023)Wu, Hu, Shi, Dziri, Suhr, Ammanabrolu, Smith, Ostendorf, and Hajishirzi}]{wu2023finegrained}
Zeqiu Wu, Yushi Hu, Weijia Shi, Nouha Dziri, Alane Suhr, Prithviraj Ammanabrolu, Noah~A. Smith, Mari Ostendorf, and Hannaneh Hajishirzi. 2023.
\newblock \href {https://openreview.net/forum?id=CSbGXyCswu} {Fine-grained human feedback gives better rewards for language model training}.
\newblock In \emph{Thirty-seventh Conference on Neural Information Processing Systems}.

\bibitem[{Xu et~al.(2024)Xu, Sun, Zheng, Geng, Zhao, Feng, Tao, Lin, and Jiang}]{xu2024wizardlm}
Can Xu, Qingfeng Sun, Kai Zheng, Xiubo Geng, Pu~Zhao, Jiazhan Feng, Chongyang Tao, Qingwei Lin, and Daxin Jiang. 2024.
\newblock \href {https://openreview.net/forum?id=CfXh93NDgH} {Wizard{LM}: Empowering large pre-trained language models to follow complex instructions}.
\newblock In \emph{The Twelfth International Conference on Learning Representations}.

\bibitem[{Yadav et~al.(2024)Yadav, Tam, Choshen, Raffel, and Bansal}]{yadav2024ties}
Prateek Yadav, Derek Tam, Leshem Choshen, Colin~A Raffel, and Mohit Bansal. 2024.
\newblock Ties-merging: Resolving interference when merging models.
\newblock \emph{Advances in Neural Information Processing Systems}, 36.

\bibitem[{Yu et~al.(2024)Yu, Jiang, Shi, YU, Liu, Zhang, Kwok, Li, Weller, and Liu}]{yu2024metamath}
Longhui Yu, Weisen Jiang, Han Shi, Jincheng YU, Zhengying Liu, Yu~Zhang, James Kwok, Zhenguo Li, Adrian Weller, and Weiyang Liu. 2024.
\newblock \href {https://openreview.net/forum?id=N8N0hgNDRt} {Metamath: Bootstrap your own mathematical questions for large language models}.
\newblock In \emph{The Twelfth International Conference on Learning Representations}.

\bibitem[{Yuan et~al.(2024)Yuan, Cui, Wang, Ding, Wang, Deng, Shan, Chen, Xie, Lin, Liu, Zhou, Peng, Liu, and Sun}]{yuan2024advancing}
Lifan Yuan, Ganqu Cui, Hanbin Wang, Ning Ding, Xingyao Wang, Jia Deng, Boji Shan, Huimin Chen, Ruobing Xie, Yankai Lin, Zhenghao Liu, Bowen Zhou, Hao Peng, Zhiyuan Liu, and Maosong Sun. 2024.
\newblock \href {https://openreview.net/forum?id=2Y1iiCqM5y} {Advancing {LLM} reasoning generalists with preference trees}.
\newblock In \emph{AI for Math Workshop @ ICML 2024}.

\bibitem[{Yue et~al.(2024{\natexlab{a}})Yue, Qu, Zhang, Fu, Huang, Sun, Su, and Chen}]{yue2024mammoth}
Xiang Yue, Xingwei Qu, Ge~Zhang, Yao Fu, Wenhao Huang, Huan Sun, Yu~Su, and Wenhu Chen. 2024{\natexlab{a}}.
\newblock \href {https://openreview.net/forum?id=yLClGs770I} {{MA}mmo{TH}: Building math generalist models through hybrid instruction tuning}.
\newblock In \emph{The Twelfth International Conference on Learning Representations}.

\bibitem[{Yue et~al.(2024{\natexlab{b}})Yue, Zheng, Zhang, and Chen}]{yue2024mammoth2}
Xiang Yue, Tuney Zheng, Ge~Zhang, and Wenhu Chen. 2024{\natexlab{b}}.
\newblock \href {https://arxiv.org/abs/2405.03548} {Mammoth2: Scaling instructions from the web}.
\newblock \emph{Preprint}, arXiv:2405.03548.

\bibitem[{Zeng et~al.(2024)Zeng, Yu, Gao, Meng, Goyal, and Chen}]{zeng2024evaluating}
Zhiyuan Zeng, Jiatong Yu, Tianyu Gao, Yu~Meng, Tanya Goyal, and Danqi Chen. 2024.
\newblock \href {https://openreview.net/forum?id=tr0KidwPLc} {Evaluating large language models at evaluating instruction following}.
\newblock In \emph{The Twelfth International Conference on Learning Representations}.

\bibitem[{Zheng et~al.(2023)Zheng, Chiang, Sheng, Zhuang, Wu, Zhuang, Lin, Li, Li, Xing, Zhang, Gonzalez, and Stoica}]{llm-as-a-judge}
Lianmin Zheng, Wei-Lin Chiang, Ying Sheng, Siyuan Zhuang, Zhanghao Wu, Yonghao Zhuang, Zi~Lin, Zhuohan Li, Dacheng Li, Eric Xing, Hao Zhang, Joseph~E Gonzalez, and Ion Stoica. 2023.
\newblock \href {https://proceedings.neurips.cc/paper_files/paper/2023/file/91f18a1287b398d378ef22505bf41832-Paper-Datasets_and_Benchmarks.pdf} {Judging llm-as-a-judge with mt-bench and chatbot arena}.
\newblock In \emph{Advances in Neural Information Processing Systems}, volume~36, pages 46595--46623. Curran Associates, Inc.

\bibitem[{Zheng et~al.(2024)Zheng, Zhang, Zhang, YeYanhan, and Luo}]{zheng-etal-2024-llamafactory}
Yaowei Zheng, Richong Zhang, Junhao Zhang, YeYanhan YeYanhan, and Zheyan Luo. 2024.
\newblock \href {https://doi.org/10.18653/v1/2024.acl-demos.38} {{L}lama{F}actory: Unified efficient fine-tuning of 100+ language models}.
\newblock In \emph{Proceedings of the 62nd Annual Meeting of the Association for Computational Linguistics (Volume 3: System Demonstrations)}, pages 400--410, Bangkok, Thailand. Association for Computational Linguistics.

\bibitem[{Zhu et~al.(2023)Zhu, Frick, Wu, Zhu, and Jiao}]{starling2023}
Banghua Zhu, Evan Frick, Tianhao Wu, Hanlin Zhu, and Jiantao Jiao. 2023.
\newblock Starling-7b: Improving llm helpfulness \& harmlessness with rlaif.

\bibitem[{Zhuo et~al.(2024)Zhuo, Vu, Chim, Hu, Yu, Widyasari, Yusuf, Zhan, He, Paul et~al.}]{zhuo2024bigcodebench}
Terry~Yue Zhuo, Minh~Chien Vu, Jenny Chim, Han Hu, Wenhao Yu, Ratnadira Widyasari, Imam Nur~Bani Yusuf, Haolan Zhan, Junda He, Indraneil Paul, et~al. 2024.
\newblock Bigcodebench: Benchmarking code generation with diverse function calls and complex instructions.
\newblock \emph{arXiv preprint arXiv:2406.15877}.

\bibitem[{Ziegler et~al.(2020)Ziegler, Stiennon, Wu, Brown, Radford, Amodei, Christiano, and Irving}]{ziegler2020finetuning}
Daniel~M. Ziegler, Nisan Stiennon, Jeffrey Wu, Tom~B. Brown, Alec Radford, Dario Amodei, Paul Christiano, and Geoffrey Irving. 2020.
\newblock \href {https://arxiv.org/abs/1909.08593} {Fine-tuning language models from human preferences}.
\newblock \emph{Preprint}, arXiv:1909.08593.

\end{thebibliography}
\bibliographystyle{acl_natbib}

\appendix

\section{Training Details}
\label{sec:appendix:training}

We use V100 GPUs for training models. We spent 2 hours training the backbone of our LLaMA-2 RM, 8 hours training our LLaMA-2 RM, and 12 hours training our Code Model. Since V100 did not support \texttt{bf16}, we adopted mixed precision training (\texttt{fp16}) for both SFT and Reward Modeling.

\subsection{Supervised Fine-Tuning (SFT)} 
We use LlamaFactory~\citep{zheng-etal-2024-llamafactory} for supervised fine-tuning (SFT). For fine-tuning the backbone of our LLaMA-2 RM, we use Alpacafarm~\citep{dubois2023alpacafarm} with a learning rate of 1e-5 and a batch size of 128. For code generation, we follow a training procedure similar to \citet{wei2024magicoder}. First, we use OSS-Instruct to fine-tune LLaMA-2-7B~\citep{touvron2023llama} for 2 epochs. Then, we continuously fine-tune the model with Magicoder-Evol-Instruct for 1 epoch. The learning rate for both stages is 1e-5, and the effective batch size is 128.

\subsection{Reward Modeling}
For reward modeling, we modify the sample code provided by TRL~\citep{vonwerra2022trl}. We trained the backbone model described in the previous section on UltraFeedback~\citep{cui2024ultrafeedback} for 1 epoch, using a learning rate of 1e-5 and a batch size of 32.

For continuous fine-tuning of LLaMA-2 RM on Auto-J Eval~\citep{li2024generative} math and code test data, we set the learning rate to 1e-6, the batch size to 8, and the number of epochs to 1.

\section{Best-of-N Sampling}
\label{sec:appendix:bon_hyperparameter}
We use vLLM~\citep{kwon2023efficient} to generate responses for reranking. For the GSM8K dataset~\citep{cobbe2021training}, we set the temperature to 1.0, top-p to 1.0, and a maximum token length of 512. In the case of MBPP~\citep{austin2021program}, we adjust the temperature to 0.1, top-p to 0.95, and maintain the same max length of 512, aligning with the hyperparameters from the \texttt{bigcode-evaluation-harness}\footnote{\url{https://github.com/bigcode-project/bigcode-evaluation-harness}} repository~\citep{zhuo2024bigcodebench}.

\section{Prompt Template}

For LLaMA-2 based models, we use the same prompt template as LLaMA-2-Chat model, as shown below:
\begin{tcolorbox}[width=\columnwidth,colback=white]
\small
\begin{verbatim}
<s>[INST] <<SYS>>
{System Prompt}
<</SYS>>

{Instruction} [/INST] {Response}<\s>
\end{verbatim}
\end{tcolorbox}

We use this template for both SFT and reward modeling. For Mistral-based models, the prompt template is modified by removing the system prompt part: 
\begin{tcolorbox}[width=\columnwidth,colback=white]
\small
\begin{verbatim}
<s>[INST] {Instruction} [/INST] {Response}<\s>
\end{verbatim}
\end{tcolorbox}

The default system prompt we used in SFT and reward modeling aligns with the original system prompt for LLaMA-2-Chat model:
\begin{tcolorbox}[width=\columnwidth,colback=white]
\small
\begin{verbatim}
You are a helpful, respectful and honest 
assistant. Always answer as helpfully as 
possible, while being safe. Your answers 
should not include any harmful, unethical,
racist, sexist, toxic, dangerous, or illegal
content. 

Please ensure that your responses are socially 
unbiased and positive in nature. If a question 
does not make any sense, or is not factually 
coherent, explain why instead of answering 
something not correct. If you don’t know the 
answer to a question, please don’t share false 
information.
\end{verbatim}
\end{tcolorbox}

The system prompt used in prompting LLaMA-2-7B-Chat for Best-of-N sampling on GSM8K is:
\begin{tcolorbox}[width=\columnwidth,colback=white]
\small
\begin{verbatim}
You are a math problem solver. Please think 
step by step and demonstrate your calculation 
steps. After your reasoning steps, you should
generate the answer by following the format 
starting with 'The answer is'
\end{verbatim}
\end{tcolorbox}

The system prompt used in prompting LLaMA-2-7B-Chat for Best-of-N sampling on MBPP is:
\begin{tcolorbox}
[width=\columnwidth,colback=white]
\small
\begin{verbatim}
Write Python code to solve the task.
\end{verbatim}
\end{tcolorbox}

\section{Dataset Details}
\label{sec:appendix:dataset}

\paragraph{Alpacafarm~\citep{dubois2023alpacafarm}}

The Alpacafarm dataset consists of 52k instructions as well as response generated by text-davinci-003 model~\citep{openai2023textdavinci003} from the original Alpaca dataset~\citep{alpaca}. Alpacafarm splits the datasets into 10k 'sft' subset for instruction fine-tuning, 10k 'pref' subset for preference learning, 20k 'unlabeled subset for training such as PPO, and 2k 'val' subset for validation. We only utilize the 10k 'sft' subset for fine-tuning the backbone of our reward model.

\paragraph{UltraFeedback~\citep{cui2024ultrafeedback}} 

This dataset consists of 64k prompts from sources including UltraChat~\citep{ding2023enhancing}, ShareGPT~\citep{vicuna2023}, Evol-Instruct~\citep{xu2024wizardlm}, TruthfulQA~\citep{lin-etal-2022-truthfulqa}, FalseQA~\citep{hu-etal-2023-wont}, and FLAN~\citep{longpre2023flan}. The responses are generated by a pool of different LLMs. The preferences are generated by GPT-4~\citep{achiam2023gpt}. In our experiment, we use a cleaned version of UltraFeedback\footnote{\url{https://huggingface.co/datasets/argilla/ultrafeedback-binarized-preferences-cleaned}}~\citep{notus2023}, which removes TruthfulQA contamination and uses the average of the preference ratings.

\paragraph{OSS-Instruct \& Magicoder Evol-Instruct~\citep{wei2024magicoder}}

OSS-Instruct consists of 75k synthesized data collected by prompting ChatGPT~\citep{achiam2023gpt} to generate a coding problem and solution based on a seed code snippet from an open-sourced platform. The Magicoder Evol-Instruct dataset, based on the work in~\citep{luo2023wizardcoder}, uses an open-source implementation~\citep{evolcode} that has been further decontaminated, resulting in 110k data points for fine-tuning. Both OSS-Instruct and Magicoder Evol-Instruct are used to fine-tune the Code Model for merging. 

\paragraph{RewardBench~\citep{lambert2024rewardbench}}

RewardBench is a benchmark designed to evaluate reward models (RMs). The datasets are categorized into core sets and prior sets. The prior sets consist of testing sets from open-sourced preference dataset such as OpenAI Summarization~\citep{openaisummarize}, Anthropic Helpful split~\citep{bai2022training}, Anthropic HHH~\citep{askell2021general}, and Stanford Human Preference (SHP)~\citep{pmlr-v162-ethayarajh22a}.

We utilize the core sets for evaluation, which include four categories: chat, chat-hard, reasoning, and safety. The chat category collects data from AlpacaEval~\citep{alpaca_eval} and MT Bench~\citep{llm-as-a-judge} to assess RMs' basic ability to discern correct responses in open-ended dialogue. Chat-Hard incorporates data from MT Bench~\citep{llm-as-a-judge} with similar ratings and LLMBar~\citep{zeng2024evaluating} data designed to challenge LLM-based judges. The reasoning category includes math data selected from PRM800K~\citep{lightman2024lets}, where the prompt is the reference answer and the rejected prompt is a wrong solution generated by GPT-4~\citep{achiam2023gpt}. The coding data utilizes HumanEvalPack~\citep{muennighoff2024octopack}, augmenting HumanEval~\citep{chen2021codex} across six programming languages, with the prompt being the reference solution and the rejected prompt being buggy solutions. Safety category comprises data from XSTest~\citep{rottger-etal-2024-xstest}, Do-Not-Answer~\citep{wang-etal-2024-answer}, and an in-development refusals dataset at AI2, aiming to accurately test models' ability to refuse dangerous content and avoid incorrect refusals triggered by similar words.

\paragraph{Auto-J Eval~\citep{li2024generative}}

Auto-J Eval's pairwise testing set includes examples from various sources: OpenAI Summarization~\citep{openaisummarize}, WebGPT~\citep{nakano2022webgpt}, Stanford SHP~\citep{pmlr-v162-ethayarajh22a}, Synthetic GPT-J~\citep{syntheticgptj}, and PKU-SafeAlignment~\citep{pkualignment}. GPT-4~\citep{achiam2023gpt} serves as the annotator. The dataset consists of categories including Summarization, Exam Questions, Code, Creative Writing, Functional Writing, Rewriting,
General Communication, and NLP Tasks. We exclude the tied examples and re-group the data into Code, Math (extract from Exam Questions category), and Others, following \citealt{yuan2024advancing}. 

\paragraph{GSM8K~\citep{cobbe2021training}}

This dataset consists of 8.5K grade school-level math problems. We use the prompt from the testing set to perform Best-of-N sampling in a zero-shot manner.

\paragraph{MBPP~\citep{austin2021program}}

This dataset consists of 1,000 crowd-sourced Python programming problems, which are entry-level problems covering standard libraries, programming, and so on. We use the testing set to perform Best-of-N sampling in a zero-shot manner.

\section{Open-Source Model Details}
\label{sec:appendix:models}

\paragraph{MetaMath~\citep{yu2024metamath}}

We use the LLaMA-2-7B based model fine-tuned by the authors for merging. The MetaMath-7B models are trained on the MetaMathQA dataset, which the authors curated by bootstrapping problems from GSM8K~\citep{cobbe2021training} and MATH~\citep{hendrycks2021measuring}. According to the original paper, the model did not trained on any data from the testing set of GSM8K and MATH.

\paragraph{MAmmoTH~\citep{yue2024mammoth}}

We merge our RM with the MAmmoTH-7B model, a LLaMA-2-7B based model fine-tuned on the MathInstruct dataset. This dataset combines a diverse range of math problems and hybrid rationales curated by the author. According to the original paper, the model did not trained on any data from the testing set of GSM8K as well as MATH.

\paragraph{Mistral-RM~\citep{mistralrm}}

We use a Mistral-based RM, initialized from \texttt{Mistral-7B-Instruct-v0.2}, trained on diverse preference datasets to evaluate our framework's adaptability. Detailed information about the training setup can be found in the author's blog.\footnote{\url{https://www.notion.so/abe03f9afdac42b9a5bee746844518d0}}

\paragraph{MAmmoTH2-Plus~\citep{yue2024mammoth2}}

To test the adaptability of our framework across different model architectures, we use the MAmmoTH2-7B-Plus and merge it with the Mistral RM. This model is fine-tuned from the MAmmoTH2-7B, which is fine-tuned from \texttt{Mistral-7B-Instruct-v0.2}, on public instruction tuning datasets to further enhance performance. According to the original paper, the model did not trained on any data from the testing set of GSM8K as well as MATH.

\section{Code Model Details}
\label{sec:appendix:code_model}
To showcase the capabilities of the fine-tuned code model, we assess its performance on two benchmarks: MBPP~\citep{austin2021program} and HumanEval~\citep{chen2021codex}, utilizing BigCodeBench~\citep{zhuo2024bigcodebench} for evaluation.
\begin{table}[t!]
    \centering
    \resizebox{0.8\columnwidth}{!}{
        \begin{tabular}{l c c}
        \toprule
        \textbf{Model} & \textbf{MBPP} & \textbf{Humaneval} \\
        \midrule
        LLaMA-2 & 18.6 & 12.2 \\
        FT on Alpacafarm & 21.0 & 15.9 \\
        Code Model & \textbf{26.2} & \textbf{31.7} \\
        \bottomrule
        \end{tabular}
    }   
    \caption{Performance on two code benchmarks.}    
    \label{tab:code_model}
\end{table}

Table~\ref{tab:code_model} demonstrates that our fine-tuned Code model outperforms the original Llama2 and the backbone of our reward model in the code benchmarks.

\section{Case Study}

\begin{table*}[pt!]
    \small
    \centering
    \begin{tabular}{|p{15cm}|}
       \hline
    {\bf Instruction (Math)}\\
    \texttt{Write a Python function `max\_fill(grid, capacity): 
import math` to solve the following problem: 
You are given a rectangular grid of wells. Each row represents a single well, 
and each 1 in a row represents a single unit of water. 
Each well has a corresponding bucket that can be used to extract water from it, 
and all buckets have the same capacity. 
Your task is to use the buckets to empty the wells.
Output the number of times you need to lower the buckets.} \\
\texttt{Example 1:} \\
\texttt{Input:} \\
\texttt{grid : [[0,0,1,0], [0,1,0,0], [1,1,1,1]] }\\
\texttt{bucket\_capacity: 1 }\\
\texttt{Output: 6 }\\
\texttt{Example 2: }\\
\texttt{Input: }\\
\texttt{grid : [[0,0,1,1], [0,0,0,0], [1,1,1,1], [0,1,1,1]] }\\
\texttt{bucket\_capacity : 2 }\\
\texttt{Output: 5 }\\
\texttt{Example 3: }\\
\texttt{Input: }\\
\texttt{grid: [[0,0,0], [0,0,0]] }\\
\texttt{bucket\_capacity : 5 }\\
\texttt{Output: 0 }\\
\texttt{Constraints: }\\
\texttt{* all wells have the same length }\\
\texttt{* 1 <= grid.length <= 10\^{}2 }\\
\texttt{* 1 <= grid[:,1].length <= 10\^{}2 }\\
\texttt{* grid[i][j] -> 0 | 1 }\\
\texttt{* 1 <= capacity <= 10}\\
{\bf Chosen Response}\\
\texttt{return sum([math.ceil(sum(arr)/capacity) for arr in grid])}\\
{\bf Rejected Response}\\
\texttt{return sum([math.floor(sum(arr)/capacity) for arr in grid])}\\
\hline\hline
{\bf Instruction (String)}\\
\texttt{Write a Python function `concatenate(strings: List[str]) -> str` to solve the following problem: }\\
\texttt{Concatenate list of strings into a single string 
>{}>{}> concatenate([]) 
'{}'{}
>{}>{}> concatenate(['a', 'b', 'c']) 
'abc'}\\
{\bf Chosen Response}\\
\texttt{return '{}'.join(strings)}\\
{\bf Rejected Response}\\
\texttt{return ' '.join(strings)}\\
\hline
       \hline
    {\bf Instruction (Bit Operation)}\\
    \texttt{Write a Python function `string\_xor(a: str, b: str) -> str` to solve the following problem:} \\
\texttt{Input are two strings a and b consisting only of 1s and 0s.} \\
\texttt{Perform binary XOR on these inputs and return result also as a string.
>{}>{}> string\_xor('010', '110') 
'100'}\\
{\bf Chosen Response}\\
\begin{verbatim}
def xor(i, j):
    if i == j:
        return '0'
    else:
        return '1'
    return ''.join(xor(x, y) for x, y in zip(a, b))
\end{verbatim}\\
{\bf Rejected Response}\\
\begin{verbatim}
def xor(i, j):
    if i == j:
        return '1'
    else:
        return '0'
    return ''.join(xor(x, y) for x, y in zip(a, b)) 
\end{verbatim}
\\
\hline
    \end{tabular}
    \caption{Passed cases after merging.}
    \label{tab:pass}
\end{table*}

\begin{table*}[pt!]
    \small
    \centering
    \begin{tabular}{|p{15cm}|}
       \hline
    {\bf Instruction (List/Sorting)}\\
    \texttt{Write a Python function `sort\_array(array)` 
to solve the following problem:} \\
\texttt{Given an array of non-negative integers, 
return a copy of the given array after sorting, 
you will sort the given array in ascending 
order if the sum( first index value, last 
index value) is odd, or sort it in descending 
order if the sum( first index value, last 
index value) is even. }\\
\texttt{Note: * don't change the given array.} \\
\texttt{Examples: }\\
\texttt{* sort\_array([]) => []}\\
\texttt{* sort\_array([5]) => [5]}\\
\texttt{* sort\_array([2, 4, 3, 0, 1, 5]) => [0, 1, 2, 3, 4, 5]} \\
\texttt{* sort\_array([2, 4, 3, 0, 1, 5, 6]) => [6, 5, 4, 3, 2, 1, 0]}\\
{\bf Chosen Response}\\
\texttt{return [] if len(array) == 0 else sorted(array, reverse= (array[0]+array[-1]) \% 2 == 0)}\\
{\bf Rejected Response}\\
\texttt{return [] if len(array) == 0 else sorted(array, reverse= (array[0]+array[-1]) \% 2 != 0)}\\
\hline\hline
{\bf Instruction (Math)}\\
\texttt{Write a Python function `derivative(xs: list)` to solve the following problem:}\\
\texttt{xs represent coefficients of a polynomial. 
xs[0] + xs[1] * x + xs[2] * x\^{}2 + ....}\\ 
\texttt{Return derivative of this polynomial in the same form.}\\
\texttt{>{}>{}> derivative([3, 1, 2, 4, 5]) 
{[}1, 4, 12, 20{]} 
>{}>{}> derivative([1, 2, 3]) 
{[}2, 6{]}}\\
{\bf Chosen Response}\\
\texttt{return [(i * x) for i, x in enumerate(xs)][1:]}\\
{\bf Rejected Response}\\
\texttt{return [(i * x) for i, x in enumerate(xs)]}\\
\hline
    \end{tabular}
    \caption{Failed cases after merging.}
    \label{tab:failed}
\end{table*}

We conducted a case study on the predictions of the reward model. We examined the changes in chosen/rejected prompt rewards on the coding subset of RewardBench before and after merging the reward models. We identified that the problems initially predicted incorrectly but corrected after merging can be categorized into math, string operations, and bit operations. The passed and failed cases are shown in Table~\ref{tab:pass} and Table~\ref{tab:failed} respectively. Of the four problems correctly predicted before merging but incorrectly predicted after merging, two were related to list/sorting operations, while the others involved mathematical operations such as modular arithmetic and derivatives.

\section{Full Results}
\label{sec:appendix:full}
Full results with different values of $\lambda$ on Best-of-N sampling and RM benchmarks are presented here.

\subsection{Best-of-N}
\label{sec:appendix:bestofn}

\begin{figure*}[hbtp]
     \centering
     \begin{subfigure}[b]{0.245\linewidth}
         \centering
         \includegraphics[width=\textwidth]{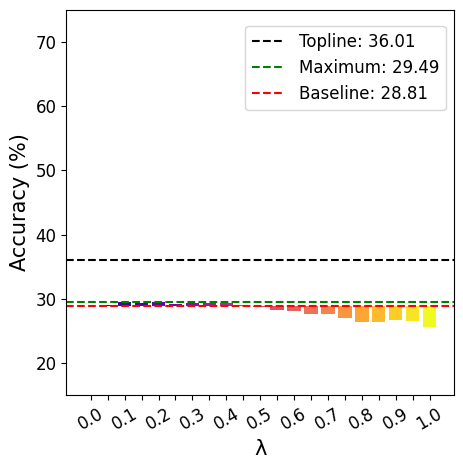}
         \caption{Best-of-2}
     \end{subfigure}
     \hfill
     \begin{subfigure}[b]{0.245\linewidth}
         \centering
         \includegraphics[width=\textwidth]{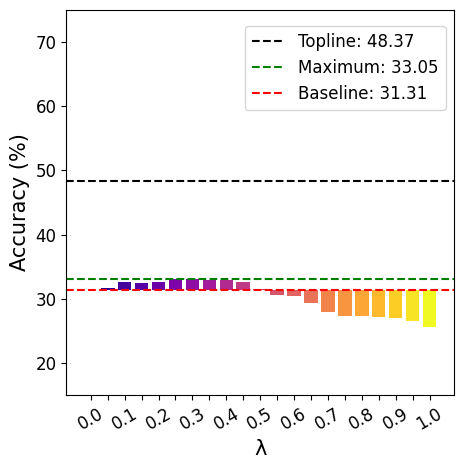}
         \caption{Best-of-4}
     \end{subfigure}
     \hfill
      \begin{subfigure}[b]{0.245\linewidth}
         \centering
         \includegraphics[width=\textwidth]{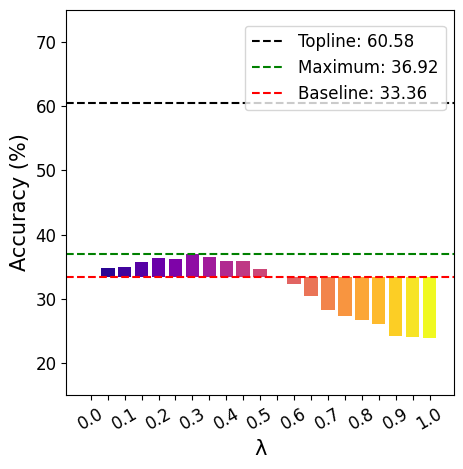}
         \caption{Best-of-8}
     \end{subfigure}
     \hfill
     \begin{subfigure}[b]{0.245\linewidth}
         \centering
         \includegraphics[width=\textwidth]{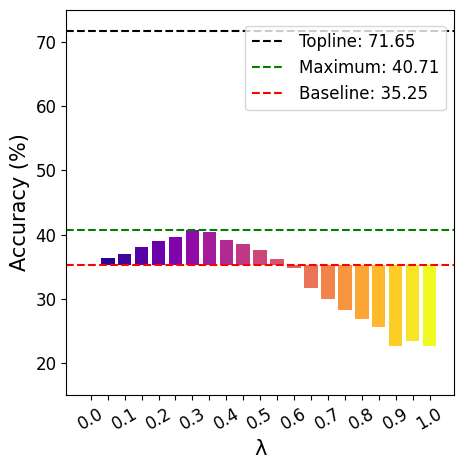}
         \caption{Best-of-16}
     \end{subfigure}
        \caption{Full results of LLaMA-2 RM + MetaMath on GSM8K.}
        \vspace{-10pt}
        \label{fig:full_bon_metamath}
\end{figure*}

\begin{figure*}[hbtp]
     \centering
     \begin{subfigure}[b]{0.245\linewidth}
         \centering
         \includegraphics[width=\textwidth]{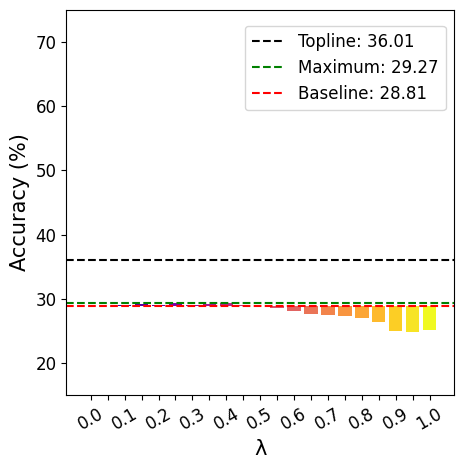}
         \caption{Best-of-2}
     \end{subfigure}
     \hfill
     \begin{subfigure}[b]{0.245\linewidth}
         \centering
         \includegraphics[width=\textwidth]{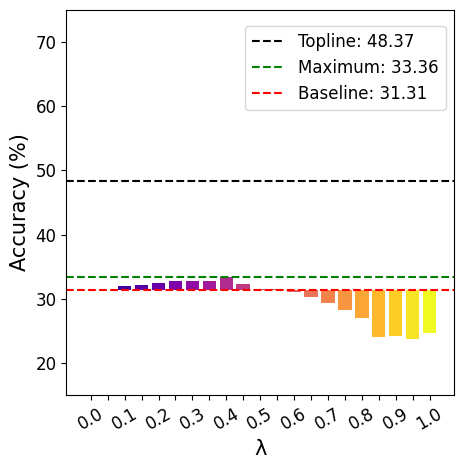}
         \caption{Best-of-4}
     \end{subfigure}
     \hfill
      \begin{subfigure}[b]{0.245\linewidth}
         \centering
         \includegraphics[width=\textwidth]{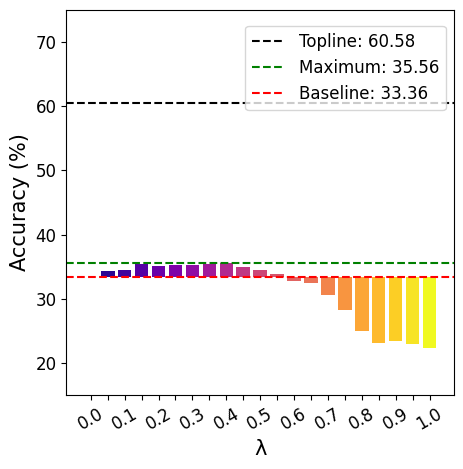}
         \caption{Best-of-8}
     \end{subfigure}
     \hfill
     \begin{subfigure}[b]{0.245\linewidth}
         \centering
         \includegraphics[width=\textwidth]{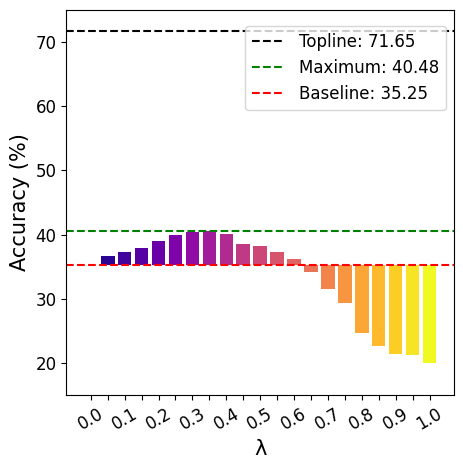}
         \caption{Best-of-16}
     \end{subfigure}
        \caption{Full results of LLaMA-2 RM + MAmmoTH on GSM8K.}
        \vspace{-10pt}
        \label{fig:full_bon_mammoth}
\end{figure*}

Figure~\ref{fig:full_bon_metamath} and ~\ref{fig:full_bon_mammoth} demonstrate the results of Best-of-N sampling on GSM8K when merging our LLaMA-2 RM with MetaMath-7B~\citep{yu2024metamath} and MAmmoTH-7B~\citep{yue2024mammoth}, respectively. DogeRM shows consistent improvement across different models being merged.

\begin{figure*}[hbtp]
     \centering
     \begin{subfigure}[b]{0.245\linewidth}
         \centering
         \includegraphics[width=\textwidth]{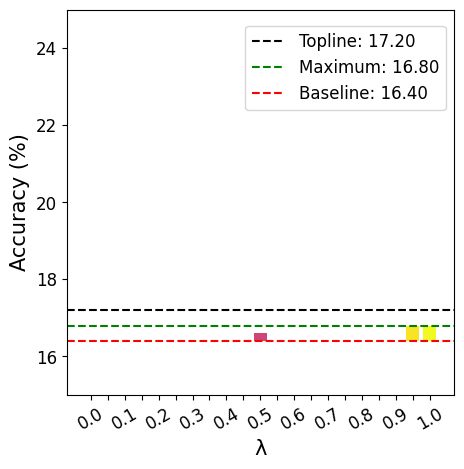}
         \caption{Best-of-2}
     \end{subfigure}
     \hfill
     \begin{subfigure}[b]{0.245\linewidth}
         \centering
         \includegraphics[width=\textwidth]{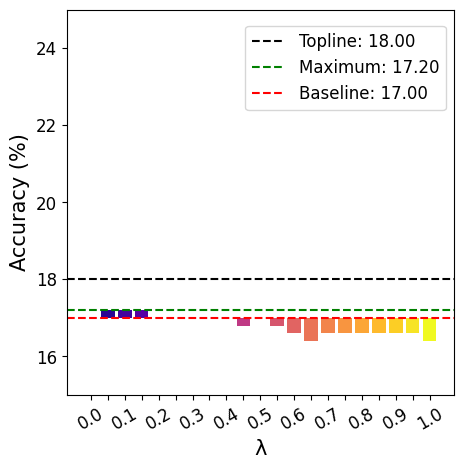}
         \caption{Best-of-4}
     \end{subfigure}
     \hfill
      \begin{subfigure}[b]{0.245\linewidth}
         \centering
         \includegraphics[width=\textwidth]{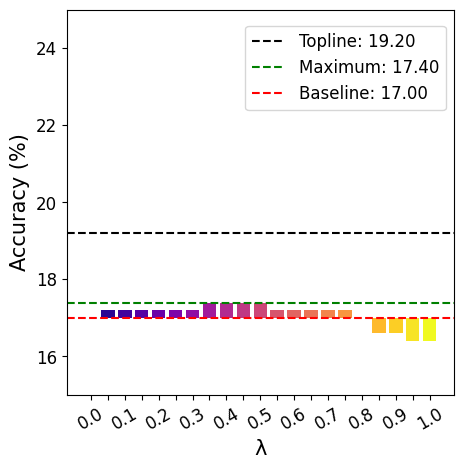}
         \caption{Best-of-8}
     \end{subfigure}
     \hfill
     \begin{subfigure}[b]{0.245\linewidth}
         \centering
         \includegraphics[width=\textwidth]{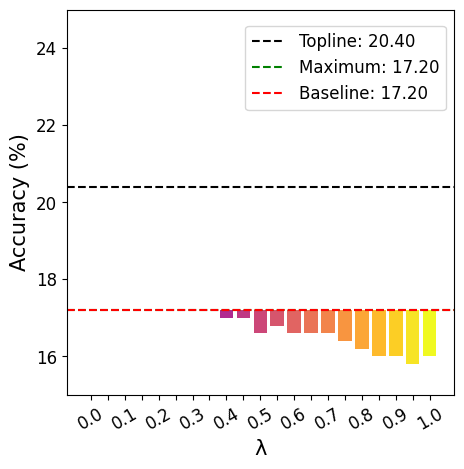}
         \caption{Best-of-16}
     \end{subfigure}
        \caption{Full results of LLaMA-2 RM + Code Model on MBPP.}
        \vspace{-10pt}
        \label{fig:full_bon_coding}
\end{figure*}

Figure~\ref{fig:full_bon_coding} 
shows the result of Best-of-N sampling on MBPP when merging our LLaMA-2 RM with the Code Model. While merging did not lead to a performance decline, the observed improvement is modest. We suspect this is attributable to the low upper bound of reranking performance (represented by the black line), which limits the potential gains from reranking in this task.

\begin{figure*}[hbtp]
     \centering
     \begin{subfigure}[b]{0.245\linewidth}
         \centering
         \includegraphics[width=\textwidth]{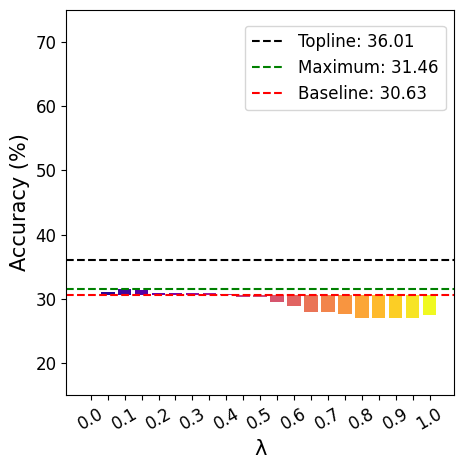}
         \caption{Best-of-2}
     \end{subfigure}
     \hfill
     \begin{subfigure}[b]{0.245\linewidth}
         \centering
         \includegraphics[width=\textwidth]{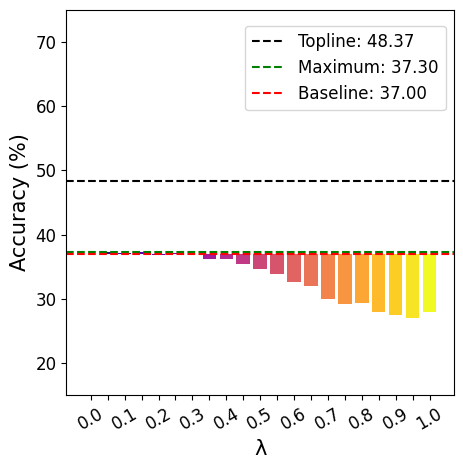}
         \caption{Best-of-4}
     \end{subfigure}
     \hfill
      \begin{subfigure}[b]{0.245\linewidth}
         \centering
         \includegraphics[width=\textwidth]{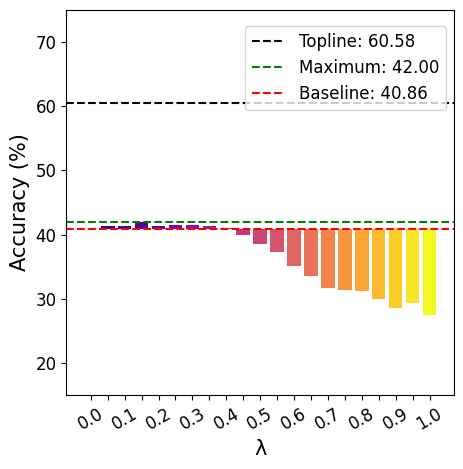}
         \caption{Best-of-8}
     \end{subfigure}
     \hfill
     \begin{subfigure}[b]{0.245\linewidth}
         \centering
         \includegraphics[width=\textwidth]{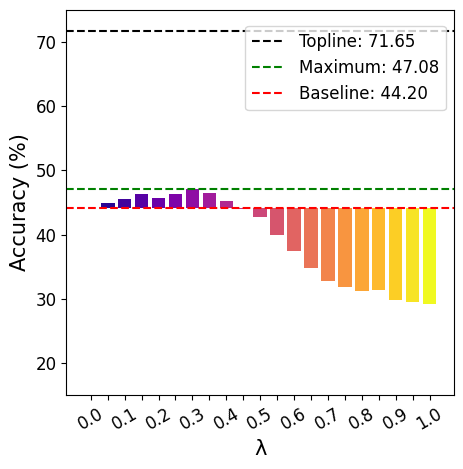}
         \caption{Best-of-16}
     \end{subfigure}
        \caption{Full results of Mistral RM + MAmmoTH2-Plus on GSM8K.}
        \vspace{-10pt}
        \label{fig:full_bon_mistral}
\end{figure*}

Finally, Figure~\ref{fig:full_bon_mistral} shows the results when merging the Mistral RM~\citep{mistralrm} with MAmmoTH2-7B-Plus~\citep{yue2024mammoth2}. DogeRM improves the reranking accuracy at an N=16 setting by 2.88\%, indicating that our method can be generalized to different model architectures.

\subsection{RewardBench}
\label{sec:appendix:rewardbench}

\begin{figure*}[hbtp]
     \centering
     \begin{subfigure}[b]{0.25\linewidth}
         \centering
         \includegraphics[width=\textwidth]{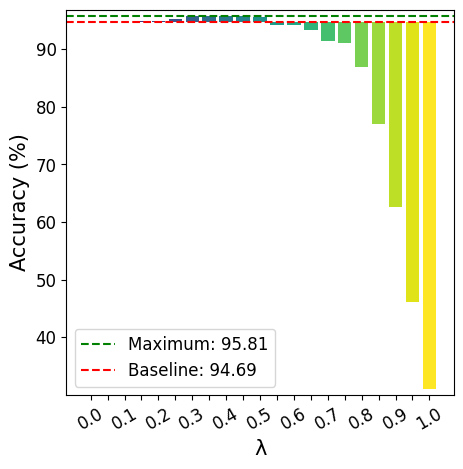}
         \caption{Chat}
     \end{subfigure}
     \begin{subfigure}[b]{0.25\linewidth}
         \centering
         \includegraphics[width=\textwidth]{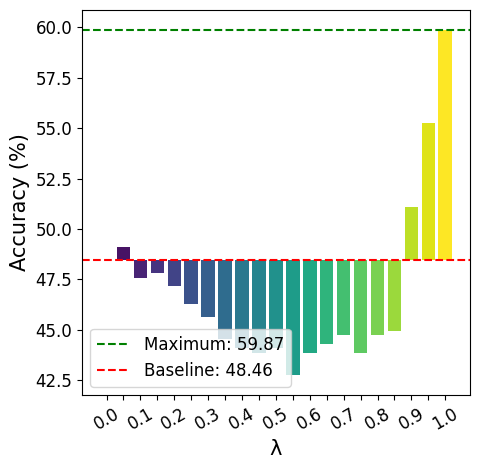}
         \caption{Chat-Hard}
     \end{subfigure}
      \begin{subfigure}[b]{0.25\linewidth}
         \centering
         \includegraphics[width=\textwidth]{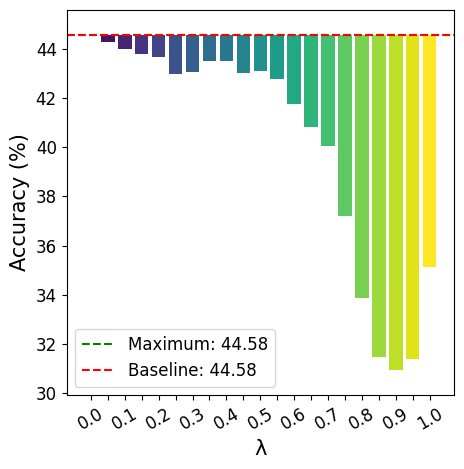}
         \caption{Safety}
     \end{subfigure}

    \vspace{0.4cm}

     \begin{subfigure}[b]{0.25\linewidth}
         \centering
         \includegraphics[width=\textwidth]{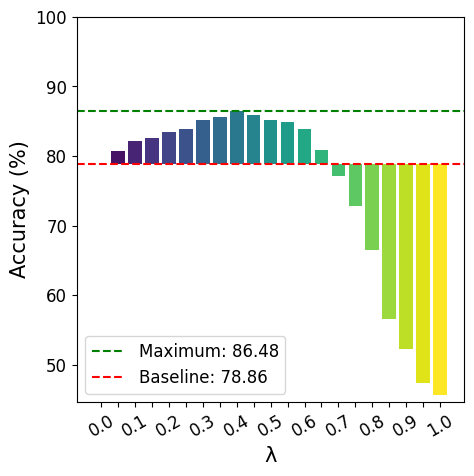}
         \caption{Code}
     \end{subfigure}
     \begin{subfigure}[b]{0.25\linewidth}
         \centering
         \includegraphics[width=\textwidth]{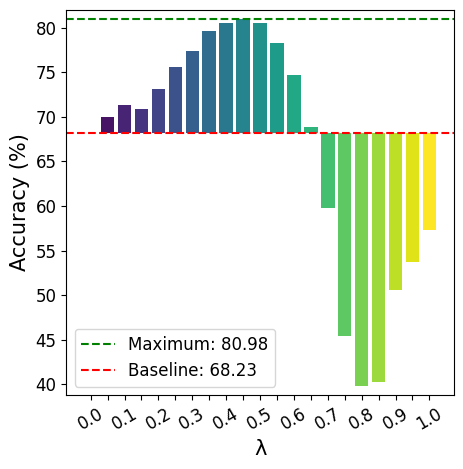}
         \caption{Math}
     \end{subfigure}
        \caption{Full results of LLaMA-2 RM + MetaMath on Reward Bench.}
        \vspace{-10pt}
        \label{fig:full_rb_metamath}
\end{figure*}

\begin{figure*}[hbtp]
     \centering
     \begin{subfigure}[b]{0.25\linewidth}
         \centering
         \includegraphics[width=\textwidth]{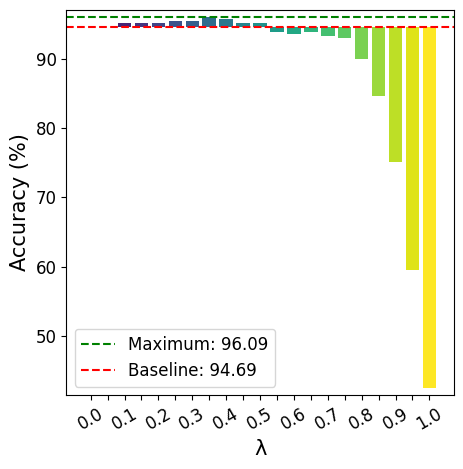}
         \caption{Chat}
     \end{subfigure}
     \begin{subfigure}[b]{0.25\linewidth}
         \centering
         \includegraphics[width=\textwidth]{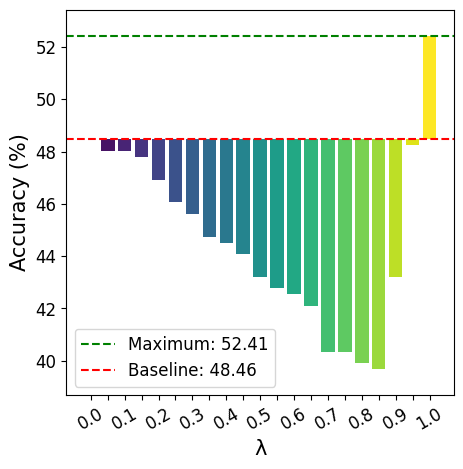}
         \caption{Chat-Hard}
     \end{subfigure}
      \begin{subfigure}[b]{0.25\linewidth}
         \centering
         \includegraphics[width=\textwidth]{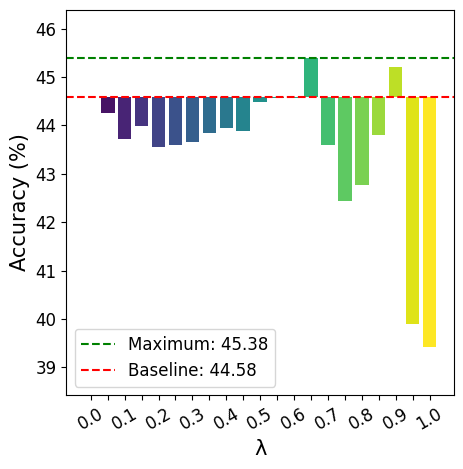}
         \caption{Safety}
     \end{subfigure}

     \vspace{0.4cm}

     \begin{subfigure}[b]{0.25\linewidth}
         \centering
         \includegraphics[width=\textwidth]{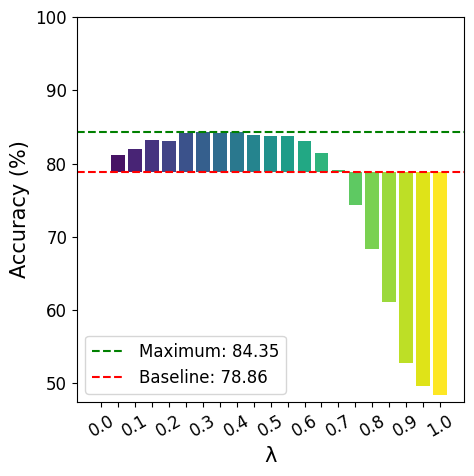}
         \caption{Code}
     \end{subfigure}
     \begin{subfigure}[b]{0.25\linewidth}
         \centering
         \includegraphics[width=\textwidth]{figures/Full_Results/RewardBench/MAmmoTH/math-prm.png}
         \caption{Math}
     \end{subfigure}
        \caption{Full results of LLaMA-2 RM + MAmmoTH on Reward Bench.}
        \vspace{-10pt}
        \label{fig:full_rb_mammoth}
\end{figure*}

Figure~\ref{fig:full_rb_metamath} and~\ref{fig:full_rb_mammoth} shows the results on different categories. We further split the reasoning category into math and coding. Merging LLaMA-2 RM with math models shows consistent improvement in both Math and Coding. The performance drop in chat-hard and safety categories can be observed. 

\begin{figure*}[hbtp]
     \centering
     \begin{subfigure}[b]{0.25\linewidth}
         \centering
         \includegraphics[width=\textwidth]{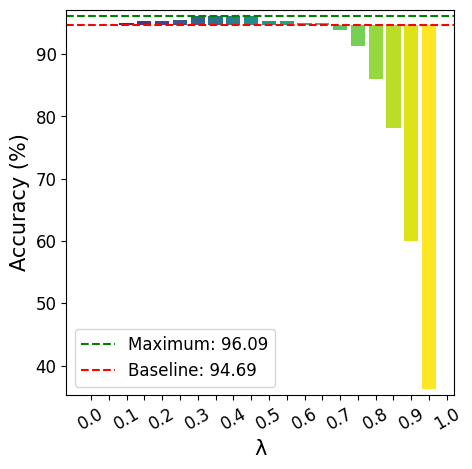}
         \caption{Chat}
     \end{subfigure}
     \begin{subfigure}[b]{0.25\linewidth}
         \centering
         \includegraphics[width=\textwidth]{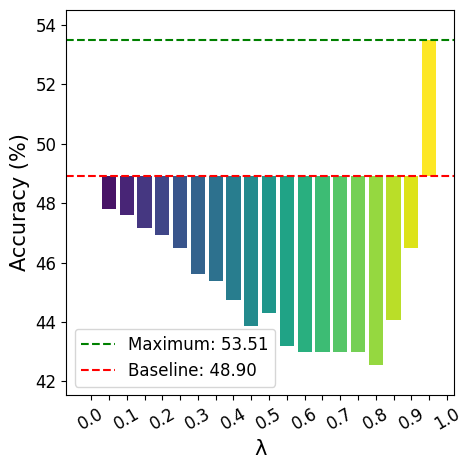}
         \caption{Chat-Hard}
     \end{subfigure}
      \begin{subfigure}[b]{0.25\linewidth}
         \centering
         \includegraphics[width=\textwidth]{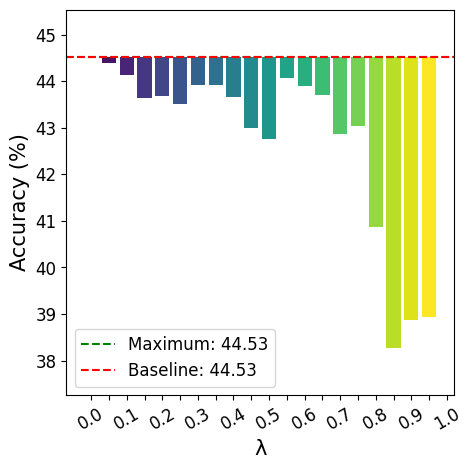}
         \caption{Safety}
     \end{subfigure}

     \vspace{0.4cm}

     \begin{subfigure}[b]{0.25\linewidth}
         \centering
         \includegraphics[width=\textwidth]{figures/Full_Results/RewardBench/Evol-Instruct/hep.png}
         \caption{Code}
     \end{subfigure}
     \begin{subfigure}[b]{0.25\linewidth}
         \centering
         \includegraphics[width=\textwidth]{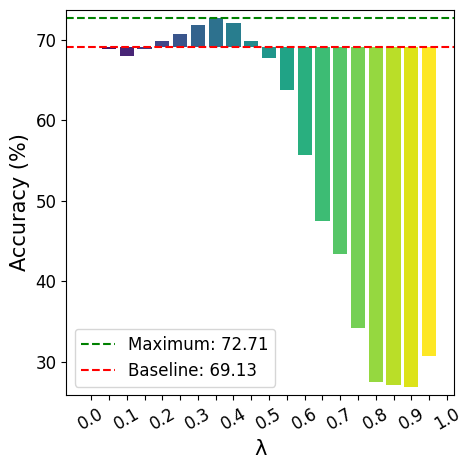}
         \caption{Math}
     \end{subfigure}
        \caption{Full results of LLaMA-2 RM + Code Model on Reward Bench.}
        \vspace{-10pt}
        \label{fig:full_rb_evol_instruct}
\end{figure*}

Figure~\ref{fig:full_rb_evol_instruct} shows the result of merging LLaMA-2 RM with the Code Model. We observe improvements in both the Math and Coding, with a performance drop in both the chat-hard and safety categories.

\begin{figure*}[hbtp]
     \centering
     \begin{subfigure}[b]{0.25\linewidth}
         \centering
         \includegraphics[width=\textwidth]{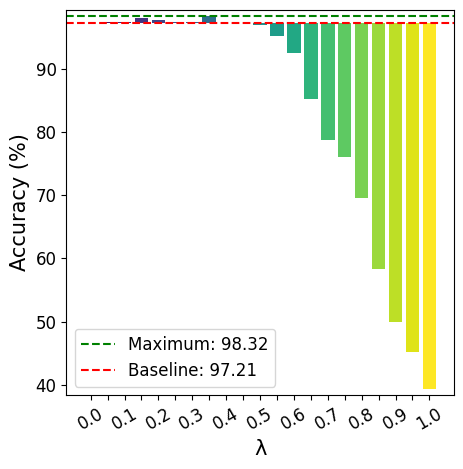}
         \caption{Chat}
     \end{subfigure}
     \begin{subfigure}[b]{0.25\linewidth}
         \centering
         \includegraphics[width=\textwidth]{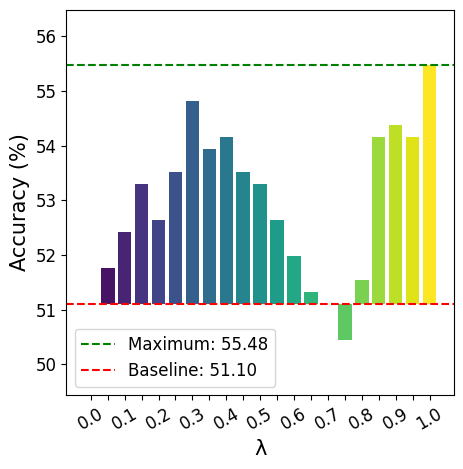}
         \caption{Chat-Hard}
     \end{subfigure}
      \begin{subfigure}[b]{0.25\linewidth}
         \centering
         \includegraphics[width=\textwidth]{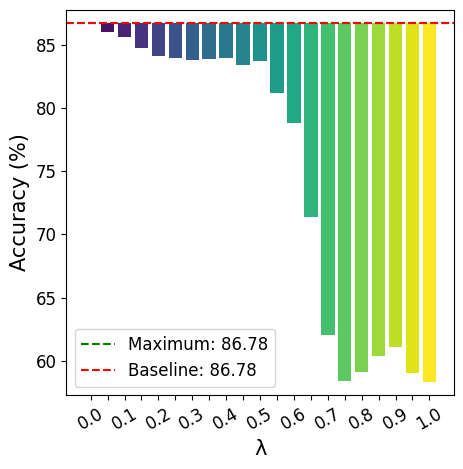}
         \caption{Safety}
     \end{subfigure}

     \vspace{0.4cm}

     \begin{subfigure}[b]{0.25\linewidth}
         \centering
         \includegraphics[width=\textwidth]{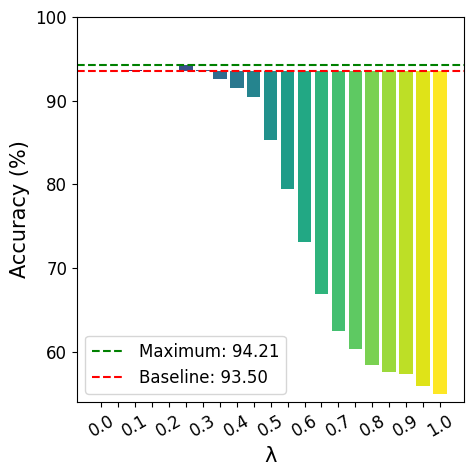}
         \caption{Code}
     \end{subfigure}
     \begin{subfigure}[b]{0.25\linewidth}
         \centering
         \includegraphics[width=\textwidth]{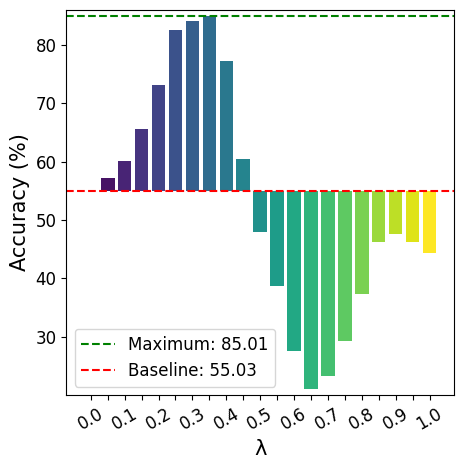}
         \caption{Math}
     \end{subfigure}
        \caption{Full results of Mistral RM + MAmmoTH2-Plus on Reward Bench.}
        \vspace{-10pt}
        \label{fig:full_rb_mistral}
\end{figure*}

Finally, Figure~\ref{fig:full_rb_mistral} shows the result of merging Mistral RM with MAmmoTH2-7B-Plus. We improve accuracy on the math subset by 30\%, while the improvement on the coding subset is minor, likely because the original RM already achieved high accuracy on this subset. An improvement in the chat-hard category can also be observed, contrary to previous cases, but a performance degradation in the safety category is found.

We believe that the performance degradation in safety aligns with observations from \citealt{yuan2024advancing}, which indicate that removing safety data from the RM training set improves reasoning performance, suggesting that modeling safety may hurt reasoning. As for the chat-hard category, we did not observe consistent performance degradation across all combinations. A deeper investigation into this is left for future work. Despite these issues, our method can effectively equip the LLaMA-2 RM with domain-specific knowledge, a finding that holds across different domains as well as different model architectures.

\subsection{Auto-J Eval}
\label{sec:appendix:autoj}

\begin{figure*}[hbtp]
     \centering
     \begin{subfigure}[b]{0.25\linewidth}
         \centering
         \includegraphics[width=\textwidth]{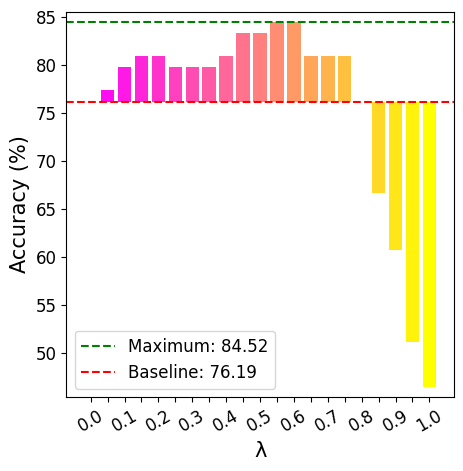}
         \caption{Code}
     \end{subfigure}
     \begin{subfigure}[b]{0.25\linewidth}
         \centering
         \includegraphics[width=\textwidth]{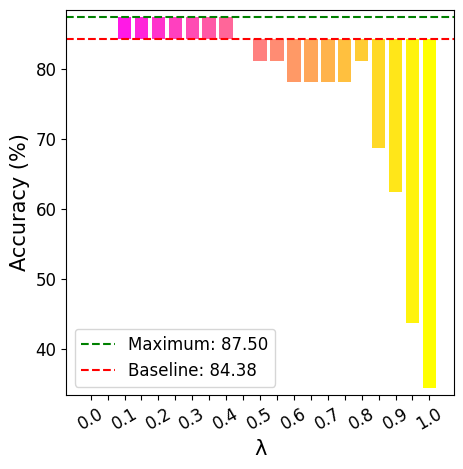}
         \caption{Math}
     \end{subfigure}
      \begin{subfigure}[b]{0.25\linewidth}
         \centering
         \includegraphics[width=\textwidth]{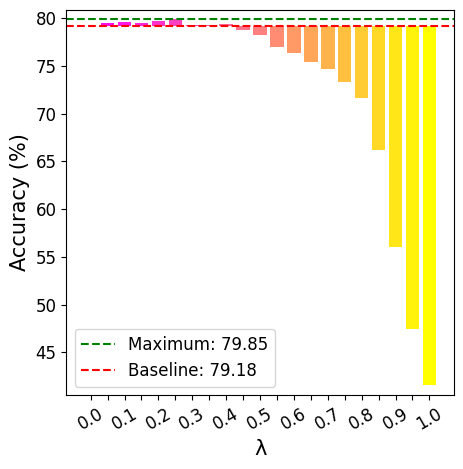}
         \caption{Others}
     \end{subfigure}
        \caption{Full results of LLaMA-2 RM + MetaMath on Auto-J Eval.}
        \vspace{-10pt}
        \label{fig:full_autoj_metamath}
\end{figure*}

\begin{figure*}[hbtp]
     \centering
     \begin{subfigure}[b]{0.25\linewidth}
         \centering
         \includegraphics[width=\textwidth]{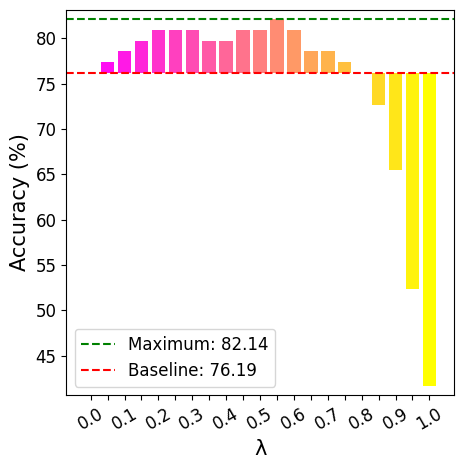}
         \caption{Code}
     \end{subfigure}
     \begin{subfigure}[b]{0.25\linewidth}
         \centering
         \includegraphics[width=\textwidth]{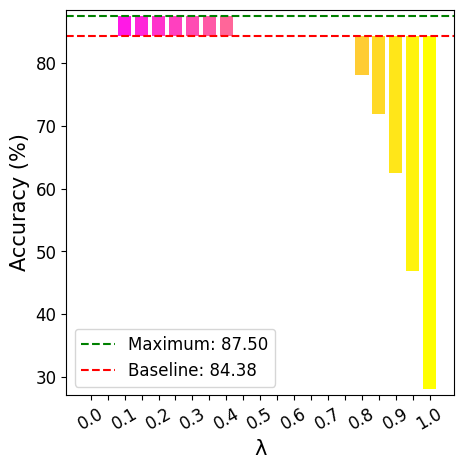}
         \caption{Math}
     \end{subfigure}
      \begin{subfigure}[b]{0.25\linewidth}
         \centering
         \includegraphics[width=\textwidth]{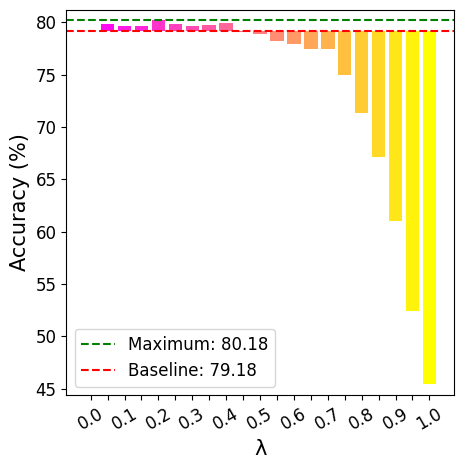}
         \caption{Others}
     \end{subfigure}
        \caption{Full results of LLaMA-2 RM + MAmmoTH on Auto-J Eval.}
        \vspace{-10pt}
        \label{fig:full_autoj_mammoth}
\end{figure*}

\begin{figure*}[hbtp]
     \centering
     \begin{subfigure}[b]{0.25\linewidth}
         \centering
         \includegraphics[width=\textwidth]{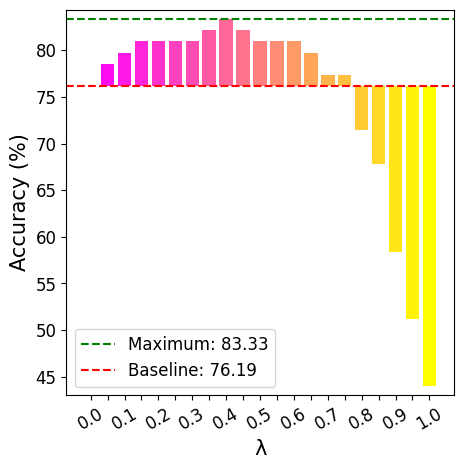}
         \caption{Code}
     \end{subfigure}
     \begin{subfigure}[b]{0.25\linewidth}
         \centering
         \includegraphics[width=\textwidth]{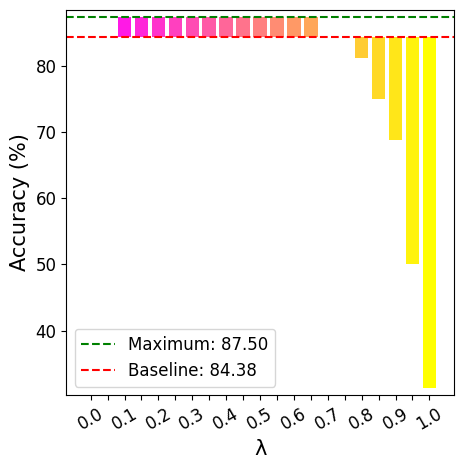}
         \caption{Math}
     \end{subfigure}
      \begin{subfigure}[b]{0.25\linewidth}
         \centering
         \includegraphics[width=\textwidth]{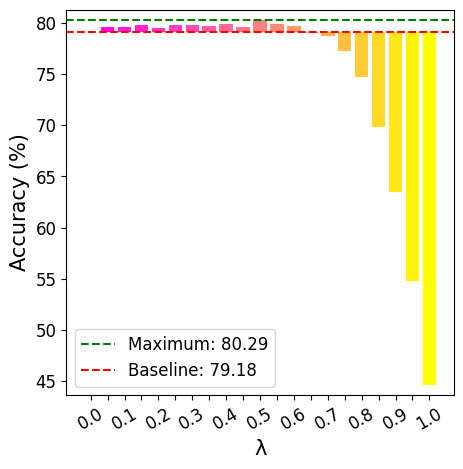}
         \caption{Others}
     \end{subfigure}
        \caption{Full results of LLaMA-2 RM + Code Model on Auto-J Eval.}
        \vspace{-10pt}
        \label{fig:full_autoj_evol_instruct}
\end{figure*}

\begin{figure*}[hbtp]
     \centering
     \begin{subfigure}[b]{0.25\linewidth}
         \centering
         \includegraphics[width=\textwidth]{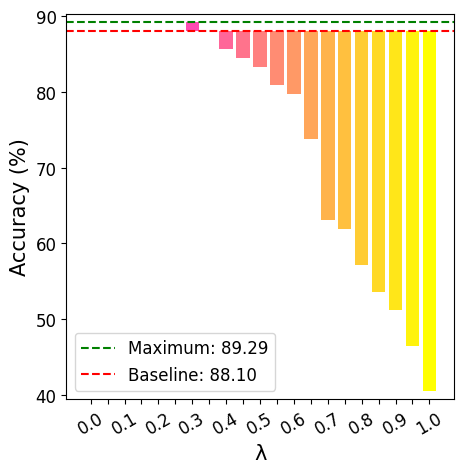}
         \caption{Code}
     \end{subfigure}
     \begin{subfigure}[b]{0.25\linewidth}
         \centering
         \includegraphics[width=\textwidth]{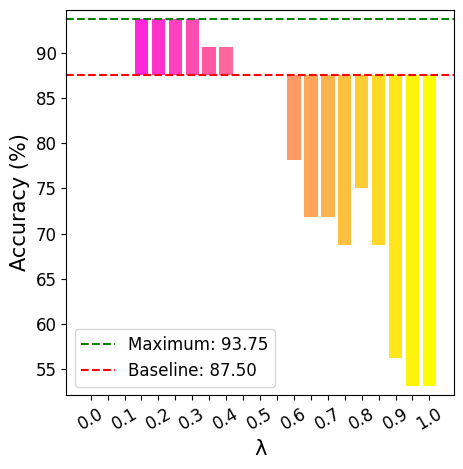}
         \caption{Math}
     \end{subfigure}
      \begin{subfigure}[b]{0.25\linewidth}
         \centering
         \includegraphics[width=\textwidth]{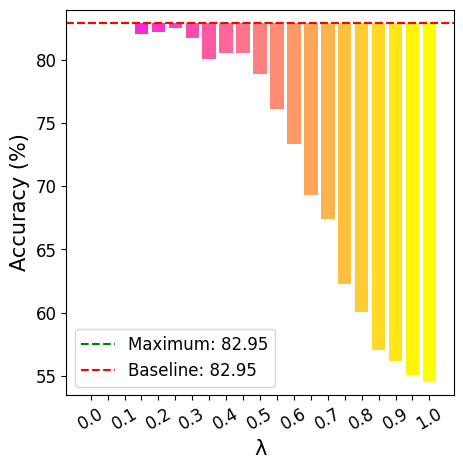}
         \caption{Others}
     \end{subfigure}
         \caption{Full results of Mistral RM + MAmmoTH2-Plus on Auto-J Eval.}
        \vspace{-10pt}
        \label{fig:full_autoj_mistral}
\end{figure*}

The results of merging LLaMA-2 RM with math models are presented in Figure~\ref{fig:full_autoj_metamath} and \ref{fig:full_autoj_mammoth}, showing improvements in both the Code and Math subsets. A similar observation can be found in Figure~\ref{fig:full_autoj_evol_instruct}, which shows the result of merging LLaMA-2 RM with the Code Model, and Figure~\ref{fig:full_autoj_mistral}, which shows the result of merging Mistral RM with MAmmoTH-2-7B-Plus. These results support the conclusion that DogeRM can equip RMs with domain-specific knowledge.

\end{document}